\documentclass[preprint,12pt]{elsarticle}
\usepackage{url} 
\usepackage{lineno}
\usepackage{soul}
\usepackage{graphicx}
\usepackage{tabularx}
\usepackage{color}
\usepackage{amsmath,amsfonts,amssymb,mathtools}
\usepackage{multirow}
\usepackage{xcolor}
\usepackage{caption}
\usepackage{subcaption}
\usepackage{float}
\usepackage{booktabs}
\usepackage{siunitx}
\usepackage{adjustbox} 
\usepackage[noabbrev]{cleveref}
\usepackage{algorithm,algpseudocode}

\begin{document}
\title{Randomized Physics-Informed Neural Networks for Bayesian Data Assimilation}

\author[UIUC]{Yifei Zong}
\author[PNNL]{David Barajas-Solano}
\author[UIUC,PNNL]{Alexandre M. Tartakovsky\corref{mycorrespondingauthor}}
\cortext[mycorrespondingauthor]{Corresponding author}
\ead{amt1998@illinois.edu}
\address[UIUC]{Department of Civil and Environmental Engineering, University of Illinois Urbana-Champaign, Urbana, IL 61801}
\address[PNNL]{Pacific Northwest National Laboratory, Richland, WA 99352}


\begin{abstract}
We propose a randomized physics-informed neural network (PINN) or rPINN method for uncertainty quantification in inverse partial differential equation (PDE) problems with noisy data. This method is used to quantify uncertainty in the inverse PDE PINN solutions. 
Recently, the Bayesian PINN (BPINN) method was proposed, where the posterior distribution of the PINN parameters was formulated using the Bayes' theorem and sampled using approximate inference methods such as the Hamiltonian Monte Carlo (HMC) and variational inference (VI) methods. In this work, we demonstrate that HMC fails to converge for non-linear inverse PDE problems. As an alternative to HMC, we sample the distribution by solving the stochastic optimization problem obtained by randomizing the PINN loss function. 
The effectiveness of the rPINN method is tested for linear and non-linear Poisson equations, and the diffusion equation with a high-dimensional space-dependent diffusion coefficient. The rPINN method provides informative distributions for all considered problems. For the linear Poisson equation, HMC and rPINN produce similar distributions, but rPINN is on average 27 times faster than HMC.  For the non-linear Poison and diffusion equations, the HMC method fails to converge because a single HMC chain cannot sample multiple modes of the posterior distribution of the PINN parameters in a reasonable amount of time.  
\end{abstract}

\maketitle


\section{Introduction}
Recent advances in deep learning have been leveraged to perform uncertainty quantification and statistical inference in complex natural and engineering systems~\cite{willard2022survey}. In particular, Bayesian Neural Networks (BNNs)~\cite{mackay1995bayesian, lampinen2001bayesian}, which are a probabilistic extension of artificial neural networks, have been widely used for stochastic model inversion and propagating model uncertainties. For example, BNNs have been employed for data-driven uncertainty analysis of river flood forecasting~\cite{boutkhamouine2020data}, rainfall-runoff modeling~\cite{khan2006bayesian}, stream flow simulations~\cite{zhang2009estimating}, and contaminant source identification~\cite{pan2021identification}. In BNNs, neural network parameters $\boldsymbol\theta$ are modeled as random variables. Given an observed dataset $\mathcal{D}$, a likelihood distribution $P(\mathcal{D}|\boldsymbol\theta)$, and a prior distribution of the parameters $P(\boldsymbol\theta)$, the Bayes' rule defines the posterior distribution of the parameters $P(\boldsymbol\theta|\mathcal{D})$, that is, the distribution of the parameters conditioned on the data. 

In many applications (e.g., subsurface flow and transport, fracture mechanics), available data is scarce and the number of parameters is very large \cite{tartakovsky2020physics}. Enforcing physics constraints can fill the gap in data to a certain degree. The physics-informed neural network (PINN) method~\cite{raissi2019physics, tartakovsky2020physics, zong2023improvedpinn} and the physics-informed conditional Karhunen-Lo\`eve expansion (PICKLE) method~\cite{tartakovsky2021pickle} are examples of so-called physics-informed machine learning methods that use the governing equations to regularize ML model training. In these methods, training is formulated as an unconstrained minimization problem over model parameters in which the loss function is the sum of a data misfit term and an approximation of a function norm of the PDE residuals (usually approximated by the $\ell^2$ norm of residuals evaluated at a subset of points over the computational domain). After adding the $\ell^2$ norm of the unknown parameters to the loss function, its minimum gives the maximum a posteriori (MAP) estimation. 

Recently, the Bayesian PINN (BPINN) method~\cite{Yang2021BPINN,psaros2023uncertainty} was introduced, where the solution of PDEs is modeled with a BNN, the standard PINN loss is treated as the negative Gaussian log-likelihood, a prior on neural network parameters is defined, and the posterior is approximated with the Markov Chain Monte Carlo (MCMC) samples or using variational inference (VI). The integration of ``physics-informed" likelihood terms in BNN training not only improves the informativeness of the posterior but also ensures that the posterior adheres with some probability to physical principles. This has shown success in COVID-19 modeling~\cite{linka2022bpinnnonlinear}, incompressible fluid flow~\cite{sun2020physics}, and seismic tomography inversion~\cite{gou2023bpinneikonal}. 

Approximating the high-dimensional posterior distributions of the PIML model parameters is challenging because the physics constraints can introduce strong correlations between parameters and cause posterior distributions to be non-Gaussian and ill-conditioned. This complexity of the posterior distributions may lead to the failure of the MCMC and VI methods. The VI method approximates the posterior distribution by optimizing a certain probabilistic metric. The mean-field VI method represents the posterior using a fully factorized (over the parameters) distribution~\cite{blei2017vi} and becomes inaccurate for posteriors with strong correlations~\cite{yao2019quality}. In the Stein variational gradient descent (SVGD) method~\cite{liu2019stein}, the posterior distribution is estimated by iteratively updating a set of samples (or ``particles'') using a transformation derived from both the gradient of the log-posterior and a kernelized repulsion term that encourages particle diversity. It was found to underestimate the covariance of high-dimensional posteriors~\cite{ba2021understanding}. MCMC methods are the gold standard for Bayesian inference because they are proven to converge to the exact posterior distributions under regularity constraints. However, they are also known to suffer from the curse of dimensionality, i.e., MCMC methods may not converge within a reasonable time. MCMC methods require a burn-in phase to ensure that the drawn samples are correctly distributed according to the target distribution, and the number of burn-in steps increases with the dimensionality and complexity of the target distribution. Furthermore, high dimensionality reduces the sampling efficiency of MCMC and increases the autocorrelation of the consecutive samples. High-dimensionality and physics constraints also increase the condition number of the posterior covariance ~\cite{zong2023randomized}, which indicates the existence of high curvature areas in the posterior and was shown to lead to biased MCMC estimates \cite{neal2011mcmc, betancourt2017conceptual}.

\begin{figure}[!htb]
    \includegraphics[width=\textwidth]{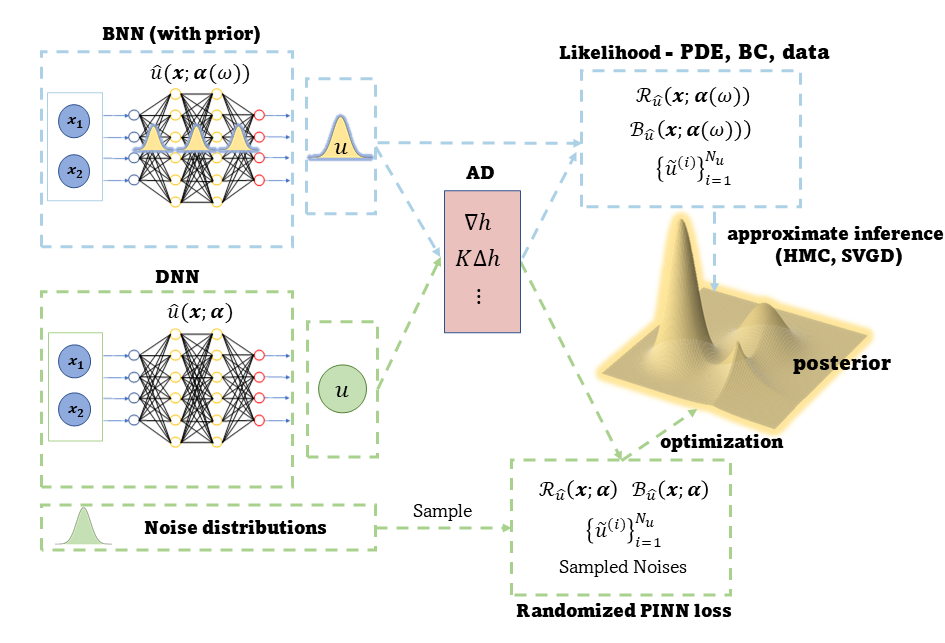}
 \caption{
 The comparison of using HMC, SVGD, and rPINN methods for sampling the posterior distribution of PINN parameters. The HMC and SVGD methods utilize the Bayesian PINN likelihood function and the neural network parameter prior (the blue dashed line). The rPINN method obtains posterior samples by solving the randomized (noise-perturbed) PINN minimization problem (the green dashed line). PDE derivatives are computed using automatic differentiation (AD).}
    \label{fig:rPINN_schematization}
\end{figure}

In this work, we propose a randomize-then-minimize approach to address the challenges in sampling posterior distributions of PINN parameters. The main idea of our randomized PINN (rPINN) approach is to add noises in the PINN loss function and approximate posterior with samples obtained as the solutions of the resulting minimization problem for different noise realizations. Figure~\ref{fig:rPINN_schematization} presents a schematic comparison of the rPINN approach and approximating BPINN posterior with MCMC and VI methods. 
Similar approaches were used to sample the distributions of the PDE-constrained optimization solutions~\cite{oliver1996conditioning}, including the randomized maximum likelihood~\cite{chen2012ensemble}, randomize-then-optimize~\cite{bardsley2014randomize}, and randomized MAP~\cite{wang2018randomized} methods. In~\cite{zong2023randomized,Wang202CMAME}, the randomized optimization approach was extended to the unconstrained minimization problem arising in the physics-informed conditional Karhunen-Lo\`eve (KL) expansion and KL-DNN methods. It was shown that the sample distribution of the randomized minimization problem converges to the exact posterior if the PDE residual is a linear function of model parameters and parameter priors are Gaussian.  

We use the No-U-Turn Sampler~\cite{hoffman2014NUTS} (NUTS), an HMC-based method, and SVGD as baselines for assessing the accuracy and efficiency of the rPINN method. We consider inverse linear and non-linear Poisson and non-linear diffusion equation problems and demonstrate that for these problems rPINN provides informative posteriors. For the linear Poisson equation, NUTS and SVGD yield posterior distributions similar to those in rPINN. However, for non-linear problems, HMC fails to converge and SVGD provides a non-informative posterior. We also consider the deep ensemble ~\cite{lakshminarayanan2017simple} method, a common method for estimating DNN model uncertainty due to random initialization of the DNN training. Our results demonstrate that the deep ensembling can significantly underestimate the total uncertainty in the PINN solutions. 


This paper is structured as follows. In Section \ref{sec:methodology}, we introduce the PINN method for solving inverse PDE problems, review the Bayesian PINN method, and formulate the rPINN method. In Section \ref{sec:results}, we compare rPINN with HMC, SVGD, and deep ensembling methods for selected inverse problems. Finally, discussions and conclusions are given in Section \ref{sec:conclusions}. The implementation codes for this work are publicly available at \url{https://github.com/geekyifei/randomized-PINN}.

\section{Methodology}\label{sec:methodology}
We consider the \emph{inverse} stationary PDE problem:
\begin{align}
 \mathcal{L}(u(\mathbf{x}),y(\mathbf{x})) &= 0 \quad \mathbf{x} \in \Omega , \label{eq:general_pde} \\
 \mathcal{B}(u(\mathbf{x})) &= g(\mathbf{x}) \quad \mathbf{x} \in \partial\Omega ,  \label{eq:general_pde_bc} 
\end{align}
where $\mathcal{L}$ is the partial differential operator, $\mathcal{B}$ is the boundary condition operator, $\Omega \subseteq \mathbb{R}^d$ ($d \in \{1, 2, 3\}$) is the simulation domain, $\partial\Omega$ is the domain boundary, $\mathbf{x} \in \Omega$ is the vector of spatial coordinates, $u(\mathbf{x})$ is the PDE state, $y(\mathbf{x})$ is the unknown space-dependent PDE parameter, and $g(\mathbf{x})$ is the unknown boundary condition function. The inverse solution is obtained using noisy measurements of $u$, $y$, and $g$.

\subsection{PINN Method for the Inverse PDE Problem}\label{sec:PINN}
In the PINN method for the inverse PDE \eqref{eq:general_pde}, the unknown PDE state variable $u(\mathbf{x})$ and PDE parameter $y(\mathbf{x})$ are represented with separate DNNs as \cite{tartakovsky2020physics}
\begin{align}\label{eq:nn_def}
u(\mathbf{x}) &\approx \hat{u}(\mathbf{x}; \boldsymbol\theta) \coloneqq \mathcal{NN}_u(\mathbf{x}; \boldsymbol\theta) \\
y(\mathbf{x}) &\approx \hat{y}(\mathbf{x}; \boldsymbol\phi) \coloneqq \mathcal{NN}_y(\mathbf{x}; \boldsymbol\phi),
\end{align}
where $\boldsymbol\theta$ and $\boldsymbol\phi$ are the parameters (weights and biases) of the corresponding DNNs. Substituting $\hat{u}(\mathbf{x}; \boldsymbol\theta)$ and $\hat{y}(\mathbf{x}; \boldsymbol\phi)$ into Eqs \eqref{eq:general_pde} and \eqref{eq:general_pde_bc} yields the PDE and boundary condition residual functions
\begin{align*}
  \mathcal{R}_L(\mathbf{x}; \boldsymbol\theta, \boldsymbol\phi) &= \mathcal{L}( \hat{u}(\mathbf{x}; \boldsymbol\theta), \hat{y}(\mathbf{x}; \boldsymbol\phi)) ,\\
  \mathcal{R}_B(\mathbf{x}; \boldsymbol\theta) &= \mathcal{B}( \hat{u}(\mathbf{x}; \boldsymbol\theta)) - g(\mathbf{x}).
\end{align*}
The DNN parameters can be estimated by minimizing the PINN loss:
\begin{align}
  \label{eq:pinn_loss_empirical}
 (\boldsymbol\theta^*, \boldsymbol\phi^*) &=  \arg\min_{\boldsymbol\theta, \boldsymbol\phi} \Big\{ \frac{\lambda_r}{N_r} \sum_{i = 1}^{N_r} \mathcal{R}_{L}(\mathbf{x}^i_r; \boldsymbol\theta, \boldsymbol\phi) ^2 + \frac{\lambda_{b}}{N_{b}} \sum_{i = 1}^{N_{b}} \mathcal{R}_{B}(\mathbf{x}^i_b; \boldsymbol\theta) ^2 \\
  &+ \frac{\lambda_y}{N_y} \sum_{i = 1}^{N_y} [\hat{y}(\mathbf{x}^i_{y}; \boldsymbol\phi) - \tilde{y}(\mathbf{x}^i_{y})]^2 +  \frac{\lambda_u}{N_u} \sum_{i = 1}^{N_u} [\hat{u}(\mathbf{x}^i_{u}; \boldsymbol\theta) - \tilde{u}(\mathbf{x}^i_{u})]^2 
 \Big\}, \nonumber
 \end{align}
where $N_{r}$ and $N_b$ are the numbers of sampling points in the simulation domain and on the boundary where the PDE and boundary residual functions are minimized, $N_{y}$ and $N_u$ are the number of assimilated noisy observations for $y$ and $u$, respectively, and $\tilde{y}(\mathbf{x}^i_{y})$ and $\tilde{u}(\mathbf{x}^i_{u})$ are the noisy observations of $y$ and $u$. The first two terms in the loss function can be viewed as "physics-informed regularization" imposing the PDE constraint on the PINN solution. The last two terms are the ``data terms'' assimilating data in the PINN solution. The weights $\lambda_r$, $\lambda_b$, $\lambda_y$, and $\lambda_u$ determine the relative importance of each term in the loss function. The PINN solution strongly depends on these weights, and the weight selection is an active area of research. In \cite{wang2021understanding}, a dynamic approach was proposed where weights are adjusted at each iteration of the minimization algorithm based on the gradient magnitude of the loss terms. In \cite{wang2022NTK}, the neural tangent kernel (NTK) based dynamic estimation of weights was presented. For computational efficiency, weights can be determined a priori. For example, an empirical relationship between weights and Péclet number was proposed for the PINN solution of advection-diffusion equation \cite{he2021physics}. In \cite{zong2023improvedpinn}, weights were determined based on the order-of-magnitude analysis of different terms in the PINN loss function. In this work, we choose a conceptually simpler ``grid-search'' approach for selecting an optimal combination of weights for the considered PDEs. 

We note that the solution of Eq~\eqref{eq:pinn_loss_empirical} is the maximum likelihood estimate (MLE) of $\boldsymbol{\theta}$ and $\boldsymbol{\phi}$. When the number of samples is small, the PINN training may require extra regularization. Adding $\ell^2$ regularization term into the loss function as~\cite{he2020pinnsubsurface}
\begin{align}\label{eq:pinn_loss_L2}
(\boldsymbol\theta^*, \boldsymbol\phi^*) &=  \arg\min_{\boldsymbol\theta, \boldsymbol\phi}  \Big\{ \frac{\lambda_r}{N_r} \sum_{i = 1}^{N_r} \mathcal{R}_{L}(\mathbf{x}^i_r; \boldsymbol\theta, \boldsymbol\phi) ^2 + \frac{\lambda_{b}}{N_{b}} \sum_{i = 1}^{N_{b}} \mathcal{R}_{B}(\mathbf{x}^i_b; \boldsymbol\theta) ^2  + \\
 &+ \frac{\lambda_y}{N_y} \sum_{i = 1}^{N_y} [\hat{y}(\mathbf{x}^i_{y}; \boldsymbol\phi) - \tilde{y}(\mathbf{x}^i_{y})]^2 +  \frac{\lambda_u}{N_u} \sum_{i = 1}^{N_u} [\hat{u}(\mathbf{x}^i_{u}; \boldsymbol\theta) - \tilde{u}(\mathbf{x}^i_{u})]^2 + \sum_{i = 1}^{N_{\theta}} \theta_i^2 + \sum_{i = 1}^{N_{\phi}} \phi_i^2\Big\} \nonumber
\end{align} 
 yields the maximum a posteriori (MAP) estimate of the PINN parameters corresponding to the independent identically distributed (i.i.d.) standard normal prior of $\boldsymbol{\theta}$ and $\boldsymbol{\phi}$. Here, $N_{\theta}$ and $N_{\phi}$ denote the dimension of $\boldsymbol{\theta}$ and $\boldsymbol{\phi}$, respectively. 
 
 In the special case of the ``additive'' parameter $y(\mathbf{x})$, where the differential operator has the form 
$\mathcal{L}(u(\mathbf{x}),y(\mathbf{x})) = \mathcal{L}_1(u(\mathbf{x})) - y(\mathbf{x})$,  we can express $y(\mathbf{x})$ in terms of the $\hat{u}$ DNN using automatic differentiation \cite{baydin2018automatic} as $y(\mathbf{x}) \approx \hat{y}(\mathbf{x},\theta) =  \mathcal{L}_1(\hat{u}(\mathbf{x},\theta) )$. This $y$ representation simplifies the problem as there is no need to introduce the second DNN, and, most importantly, the first term in the loss functions in Eqs \eqref{eq:pinn_loss_empirical} and \eqref{eq:pinn_loss_L2} drops out because $ \mathcal{R}_L(\mathbf{x}; \boldsymbol\theta) = 0$ for all $\mathbf{x}$ and $\boldsymbol\theta$.

\subsection{Bayesian Estimate of Uncertainty in the Inverse PINN Solution}\label{sec:BPINN}
The uncertainty in the estimates of $\boldsymbol\theta$ and $\boldsymbol\phi$ (and the corresponding $\hat{u}(\mathbf{x};\boldsymbol\theta)$ and $\hat{y}(\mathbf{x};\boldsymbol\phi)$) can be quantified using the Bayesian framework, which defines the posterior distribution of DNN parameters conditioned on the measurements $\mathcal{D}$ as
\begin{equation}\label{eq:posterior}
P(\boldsymbol\theta, \boldsymbol\phi | \mathcal{D}) = \frac{1}{Z(\mathcal{D})} P(\mathcal{D}| \boldsymbol\theta, \boldsymbol\phi )P(\boldsymbol\theta )P(\boldsymbol\phi ), \quad Z(\mathcal{D}) \coloneqq \int P(\mathcal{D}| \boldsymbol\theta, \boldsymbol\phi )P(\boldsymbol\theta)P(\boldsymbol\phi ) \, d\boldsymbol\theta d\boldsymbol\phi,
\end{equation}
where  $P(\mathcal{D} | \boldsymbol\theta, \boldsymbol\phi)$ is the likelihood function, $P(\boldsymbol\theta )$ and $P(\boldsymbol\phi )$ are the prior distributions of  $\boldsymbol\theta$ and $\boldsymbol\phi$, and $Z(\mathcal{D})$ is the so-called ``marginal likelihood.'' 
For the considered PINN formulation, $\mathcal{D}$ includes $\mathcal{D}_{r} = \{ \mathcal{R}_L(\mathbf{x}^i_r; \boldsymbol\theta, \boldsymbol\phi) \}_{i = 1}^{N_{r}} $ and $
  \mathcal{D}_{b} = \{ \mathcal{R}_B(\mathbf{x}^i_b; \boldsymbol\theta) \}_{i = 1}^{N_{b}} $ (we assume that the residuals at the sampling points are independently and normally distributed around zero), and $\mathcal{D}_{y} = \{ \tilde{y}(\mathbf{x}^i_y) \}_{i = 1}^{N_{y}}$ and $\mathcal{D}_{u} = \{ \tilde{u}(\mathbf{x}^i_u) \}_{i = 1}^{N_{u}}$, which are the noisy measurements of $y$ and $u$, respectively.  

Following the Bayesian PINN formulation of \cite{sun2020physics, Yang2021BPINN, psaros2023uncertainty},  the likelihood function takes the form:
\begin{align}\label{eq:likelihood}
    P(\mathcal{D} | \boldsymbol\theta, \boldsymbol\phi) &= P(\mathcal{D}_{r} | \boldsymbol\theta, \boldsymbol\phi)P(\mathcal{D}_{b} | \boldsymbol\theta)P(\mathcal{D}_{y} | \boldsymbol\phi)P(\mathcal{D}_{u} | \boldsymbol\theta),
\end{align}
where
\begin{align}\label{eq:bayes_likelihood}
    P(\mathcal{D}_{r} | \boldsymbol\theta, \boldsymbol\phi) &= \prod_i^{N_{r}}\frac{1}{\sqrt{2\pi\sigma_{r}^2}}\exp \Big\{ -\frac{\mathcal{R}_{L}(\mathbf{x}^i_r; \boldsymbol\theta, \boldsymbol\phi) ^2}{2\sigma_{r}^2} \Big\} , \\
    P(\mathcal{D}_{b} | \boldsymbol\theta) &= \prod_i^{N_b}\frac{1}{\sqrt{2\pi\sigma_{b}^2}}\exp \Big\{ -\frac{\mathcal{R}_{B}(\mathbf{x}^i_b; \boldsymbol\theta) ^2}{2\sigma_{b}^2} \Big\}  , \\
    P(\mathcal{D}_{y} | \boldsymbol\phi) &= \prod_i^{N_y}\frac{1}{\sqrt{2\pi\sigma_{y}^2}}\exp \Big\{ -\frac{[\hat{y}(\mathbf{x}^i_{y}; \boldsymbol\phi) - \tilde{y}(\mathbf{x}^i_{y})]^2}{2\sigma_{y}^2} \Big\}  , \\
    P(\mathcal{D}_{u} | \boldsymbol\theta) &= \prod_i^{N_u}\frac{1}{\sqrt{2\pi\sigma_{u}^2}}\exp \Big\{ -\frac{[\hat{u}(\mathbf{x}^i_{u}; \boldsymbol\theta) - \tilde{u}(\mathbf{x}^i_{u})]^2}{2\sigma_{u}^2} \Big\} .
\end{align}
Here, $\sigma_r$, $\sigma_b$, $\sigma_y$, $\sigma_u$ are the Bayesian model hyperparameters (standard deviations in the Gaussian likelihood). We will later propose relationships between these values and the noise variances in the measurements of $u$, $y$, and $g$.

The prior distribution of $\boldsymbol{\theta}$ and $\boldsymbol{\phi}$ corresponding to the $\ell^2$ regularization terms in the PINN loss function in Eq \eqref{eq:pinn_loss_L2} is isotropic normal:
\begin{align}\label{eq:prior}
    P(\boldsymbol\theta) &= \prod_{i=1}^{N_\theta}\frac{1}{ \sqrt{2\pi\sigma^2_p}} \exp \left (-\frac{\theta_i^2}{2\sigma_p^2} \right ), \\
    P(\boldsymbol\phi) &= \prod_{i=1}^{N_\phi}\frac{1}{ \sqrt{2\pi\sigma^2_p}} \exp \left (-\frac{\phi_i^2}{2\sigma_p^2} \right ),
\end{align}
where $\sigma_p$ is the prior standard deviation.

Due to the intractability of evaluating the high-dimensional integral defining the marginal likelihood $Z(\boldsymbol\theta)$, approximate inference methods such as HMC and VI were used to estimate the posterior of the PINN parameters. 

Any posterior distribution can be understood as the canonical distribution corresponding to an energy function $E(\boldsymbol\theta, \boldsymbol\phi)$ ~\cite{neal2011mcmc}, that is,
\begin{eqnarray}
    P(\boldsymbol\theta, \boldsymbol\phi | \mathcal{D})= \frac{1}{Z}\exp(-E(\boldsymbol\theta, \boldsymbol\phi)),
\end{eqnarray}
Here, the energy function takes the form
\begin{align}\label{eq:energy_loss}
E(\boldsymbol\theta, \boldsymbol\phi) &=  \frac{1}{2\sigma_{r}^2} \sum_{i = 1}^{N_r} \mathcal{R}_{L}(\mathbf{x}^i_r; \boldsymbol\theta, \boldsymbol\phi) ^2 + \frac{1}{2\sigma_{b}^2} \sum_{i = 1}^{N_{b}} \mathcal{R}_{B}(\mathbf{x}^i_b; \boldsymbol\theta) ^2  + \\
 &+ \frac{1}{2\sigma_{y}^2} \sum_{i = 1}^{N_y} [\hat{y}(\mathbf{x}^i_{y}; \boldsymbol\phi) - \tilde{y}(\mathbf{x}^i_{y})]^2 +  \frac{1}{2\sigma_{u}^2}  \sum_{i = 1}^{N_u} [\hat{u}(\mathbf{x}^i_{u}; \boldsymbol\theta) - \tilde{u}(\mathbf{x}^i_{u})]^2 + \frac{1}{2\sigma_{p}^2}  \sum_{i = 1}^{N_{\theta}} \theta_i^2 + \frac{1}{2\sigma_{p}^2} \sum_{i = 1}^{N_{\phi}} \phi_i^2 . \nonumber
\end{align} 
We note that the mode of the posterior (or,  the MAP) is 
$$
(\boldsymbol\theta^*,\boldsymbol\phi^*) = 
\max_{\boldsymbol\theta,\boldsymbol\phi}
P(\boldsymbol\theta, \boldsymbol\phi | \mathcal{D})
=
\min_{\boldsymbol\theta,\boldsymbol\phi} E(\boldsymbol\theta, \boldsymbol\phi).
$$
In the following, we establish relationships between $\sigma^2_r$, $\sigma^2_b$, $\sigma^2_y$, $\sigma^2_u$, and $\sigma^2_p$ and the weights $\lambda_r$, $\lambda_b$, $\lambda_y$, and $\lambda_u$ in the PINN loss function \eqref{eq:pinn_loss_L2} by requiring the maximum of $P(\boldsymbol\theta, \boldsymbol\phi | \mathcal{D})$ to be equal to the MAP estimate given by PINN.

The PINN loss function can be multiplied by $1/2 \sigma^2_p$ as
\begin{eqnarray}\label{eq:pinn_loss_norm}
&&L(\boldsymbol\theta, \boldsymbol\phi) = \frac{\lambda_r}{2N_r \sigma^2_p } \sum_{i = 1}^{N_r} \mathcal{R}_{L}(\mathbf{x}^i_r; \boldsymbol\theta, \boldsymbol\phi) ^2 + \frac{\lambda_{b}}{2N_b\sigma^2_p} \sum_{i = 1}^{N_{b}} \mathcal{R}_{B}(\mathbf{x}^i_b; \boldsymbol\theta) ^2  + \\
 &+& \frac{\lambda_y}{2N_y \sigma^2_p} \sum_{i = 1}^{N_y} [\hat{y}(\mathbf{x}^i_{y}; \boldsymbol\phi) - \tilde{y}(\mathbf{x}^i_{y})]^2 +  \frac{\lambda_u}{2N_u \sigma^2_p} \sum_{i = 1}^{N_u} [\hat{u}(\mathbf{x}^i_{u}; \boldsymbol\theta) - \tilde{u}(\mathbf{x}^i_{u})]^2 + \frac{1}{2 \sigma^2_p} \sum_{i = 1}^{N_{\theta}} \theta_i^2 + \frac{1}{2 \sigma^2_p } \sum_{i = 1}^{N_{\phi}} \phi_i^2  \nonumber
\end{eqnarray} 
without changing its minimum (and the PINN MAP estimate). Then, setting $L(\boldsymbol\theta, \boldsymbol\phi)  =     E(\boldsymbol\theta, \boldsymbol\phi)$ we obtain
\begin{eqnarray}\label{eq:likelihood_variances}
    \sigma^2_r = \frac{N_r \sigma^2_p}{\lambda_r}, \quad
    \sigma^2_b = \frac{N_b \sigma^2_p}{\lambda_b}, \quad
    \sigma^2_y = \frac{N_y \sigma^2_p}{\lambda_y}, \quad
    \sigma^2_u = \frac{N_u \sigma^2_p}{\lambda_u}. 
\end{eqnarray}
 
The posterior distribution of $\boldsymbol\theta$ and $\boldsymbol\phi$ might strongly depend on the prior variance $\sigma^2_p$. Usually, prior distributions are selected based on pre-existing knowledge of the system, e.g., based on the observations of $y$ and $u$. However, in the PINN model, the prior distributions must be imposed on the parameters in the DNN models, and a priori knowledge about $u$ and $y$ cannot be easily transferred to the DNN parameters because of the non-linearity of the DNNs.


Here, we propose an empirical relationship between $\sigma^2_p$ and the measurement errors. Following  \cite{Yang2021BPINN}, we assume that the measurement and residual errors have the same variance $\sigma^2$. Next, we require that $\sigma^2_y \le \sigma^2$ and $\sigma^2_u \le \sigma^2$. In the considered later examples, we find that $\lambda_y \le \lambda_u$, and we assumed that $N_u \le N_y$. For this combination of the parameters, we propose the following model for $\sigma^2_p$:
\begin{equation}\label{eq:sigmap}
\sigma^2_p = \sigma^2 \frac{\lambda_y}{N_y}.    
\end{equation}
Substituting this expression in Eq \eqref{eq:likelihood_variances} yields
\begin{eqnarray}\label{eq:rpinn_weights}
    \sigma^2_r = \sigma^2 \frac{\lambda_y N_r}{\lambda_r N_y}, \quad
    \sigma^2_b = \sigma^2 \frac{\lambda_y N_b}{\lambda_b N_y}, \quad
    \sigma^2_y = \sigma^2, \quad
    \sigma^2_u = \sigma^2 \frac{\lambda_y N_u}{\lambda_u N_y}. 
\end{eqnarray}
We note that it is also possible to treat $\sigma^2_p$ as a hyper-parameter and select it to maximize the log predictive probability (LPP) and/or minimize the difference between the MAP estimate given by Eq \eqref{eq:pinn_loss_norm} and the reference solution. The LPP of $u$ with respect to the known values of $u$ is defined as~\cite{rasmussen2006gaussian}:
\begin{eqnarray}\label{eq:lpp}
    \mathrm{LPP} = -\sum_{i = 1}^{N} \left\{ \frac{[\mu_u (\mathbf{x}_i | \mathcal{D}) - u(\mathbf{x}_i)]^2}{2\sigma_u^2( \mathbf{x}_i | \mathcal{D})} +  \frac{1}{2}\log [2\pi \sigma_u^2(\mathbf{x}_i | \mathcal{D})] \right\},
\end{eqnarray}
where the summation is over all points (elements) with known reference solutions. 
If the reference solution is not available, the cross-validation \cite{rasmussen2006gaussian}, meta-learning \cite{feurer2015initializing}, or empirical Bayes \cite{murphy2022probabilistic} approaches can be used instead to determine the value of $\sigma^2_p$. 

For the general case when different variables have different measurement errors, our approach can be generalized as: 
\begin{itemize}
\item set $\lambda_y=1 / \tilde{\sigma}^{2}_y$ and $\lambda_u=1 / \tilde{\sigma}^{2}_u$, where  $\tilde{\sigma}^{2}_y$ and $\tilde{\sigma}^{2}_u$ are the variances of the $y$ and $u$ measurement errors, respectively,
\item for the uncertain boundary condition function $g(x)$, set $\lambda_b= 1 / 
 \tilde{\sigma}^{2}_b$, where  $\tilde{\sigma}^{2}_b$ is the variances of $g(x)$ measurement errors,
\item determine the remaining weights (e.g., $\lambda_r$) in the PINN loss function in Eq \eqref{eq:pinn_loss_norm} such as to minimize the difference between the PINN solution and the reference or using the leave-one-out approach if the reference is not available,
\item evaluate the likelihood parameters using Eq \eqref{eq:likelihood_variances}.  
\end{itemize}

We note that in the BPINN model in \cite{Yang2021BPINN}, the parameters in the likelihood were chosen as:
\begin{eqnarray}\label{eq:likelihood_variances_uniform}
    \sigma^2_r =
    \sigma^2_b = 
    \sigma^2_y = 
    \sigma^2_u = \sigma^2.  
\end{eqnarray}
and the prior of the DNN parameters is assumed to be the independent standard Gaussian distribution.  Our results show that this choice of  $\sigma^2_r$, $\sigma^2_b$, $\sigma^2_y$, and $\sigma^2_u$, which is independent of the number of measurements and residual points, might lead to non-informative priors and inconsistent MAP estimates. 
For example, for the Poisson equation considered in Section \ref{sec:linear_poisson}, this choice of parameters leads to larger MAP predictive errors (the difference between the MAP estimate and the reference solution) and smaller values of the log predictive probability (the more informative posterior corresponds to larger LPP) with the increasing number of measurements. 

To understand why parameters in Eq \eqref{eq:likelihood_variances_uniform}  might lead to inconsistent prior estimates, we note that the corresponding MAP estimate is given by the values of $\boldsymbol\theta$ and $\boldsymbol\phi$ that minimize the loss function 
\begin{eqnarray}\label{eq:pinn_loss_Karn}
&&L(\boldsymbol\theta, \boldsymbol\phi) = \frac{1}{\sigma^2 } \sum_{i = 1}^{N_r} \mathcal{R}_{L}(\mathbf{x}^i_r; \boldsymbol\theta, \boldsymbol\phi) ^2 + \frac{1}{\sigma^2} \sum_{i = 1}^{N_{b}} \mathcal{R}_{B}(\mathbf{x}^i_b; \boldsymbol\theta) ^2  + \\
 &+& \frac{1}{ \sigma^2} \sum_{i = 1}^{N_y} [\hat{y}(\mathbf{x}^i_{y}; \boldsymbol\phi) - \tilde{y}(\mathbf{x}^i_{y})]^2 +  \frac{1}{\sigma^2} \sum_{i = 1}^{N_u} [\hat{u}(\mathbf{x}^i_{u}; \boldsymbol\theta) - \tilde{u}(\mathbf{x}^i_{u})]^2 + \frac{1}{ \sigma^2_p} \sum_{i = 1}^{N_{\theta}} \theta_i^2 + \frac{1}{\sigma^2_p} \sum_{i = 1}^{N_{\phi}} \phi_i^2.  \nonumber
\end{eqnarray} 
This particular choice of weights in the PINN loss function can provide biased PINN (MAP) estimates as stated in Section \ref{sec:PINN}. 

\subsection{Randomized PINN: Optimization-based Posterior Approximation}\label{sec:rpinn}

The randomized PINN method is formulated by injecting random noises in the PINN loss function \eqref{eq:pinn_loss_norm} (or, the BPINN energy function \eqref{eq:energy_loss}) as:
\begin{align}\label{eq:rPINN_loss}
 &L^r (\boldsymbol\theta, \boldsymbol\phi; \boldsymbol\omega_r, \boldsymbol\omega_b, \boldsymbol\omega_y, \boldsymbol\omega_u, \boldsymbol\omega_\theta, \boldsymbol\omega_\phi)  
  = \frac{1}{2\sigma_{r}^2} \sum_{i = 1}^{N_r} [ \mathcal{R}_{L}(\mathbf{x}^i_r; \boldsymbol\theta, \boldsymbol\phi) - \omega_r^i]^2 + \frac{1}{2\sigma_{b}^2} \sum_{i = 1}^{N_{b}} [ \mathcal{R}_{B}(\mathbf{x}^i_b; \boldsymbol\theta) - \omega_b^i ]^2 \nonumber \\ 
 &+ \frac{1}{2\sigma_{y}^2} \sum_{i = 1}^{N_y} [\hat{y}(\mathbf{x}^i_{y}; \boldsymbol\phi) - \tilde{y}(\mathbf{x}^i_{y}) - \omega_y^i]^2 +  \frac{1}{2\sigma_{u}^2}  \sum_{i = 1}^{N_u} [\hat{u}(\mathbf{x}^i_{u}; \boldsymbol\theta) - \tilde{u}(\mathbf{x}^i_{u}) - \omega_u^i]^2 \nonumber \\
 & + \frac{1}{2\sigma_{p}^2}  \sum_{i = 1}^{N_{\theta}} [\theta_i - \omega_\theta^i]^2 + \frac{1}{2\sigma_{p}^2} \sum_{i = 1}^{N_{\phi}} [\phi_i - \omega_\phi^i]^2,
\end{align}
where the injected noise terms have the following normal distributions: $\boldsymbol\omega_r \sim \mathcal{N}(0, \sigma^2_r\mathbf{I})$, $\boldsymbol\omega_b \sim \mathcal{N}(0, \sigma^2_b\mathbf{I})$, $\boldsymbol\omega_y \sim \mathcal{N}(0, \sigma^2_y\mathbf{I})$, $\boldsymbol\omega_u \sim \mathcal{N}(0, \sigma^2_u\mathbf{I})$, $\boldsymbol\omega_\theta \sim \mathcal{N}(0, \sigma^2_p\mathbf{I})$, $\boldsymbol\omega_\phi\sim \mathcal{N}(0, \sigma^2_p\mathbf{I})$. Here, the variances of the injected random noise distributions are given by the inverse of the weights of the corresponding terms in the loss function. 

In rPINN, the posterior distribution $P(\boldsymbol\theta, \boldsymbol\phi | \mathcal{D})$ is approximated by samples obtained by minimizing the randomized loss function for independent samples of the noise terms according to Algorithm \ref{alg:randomized_pinn}. This algorithm can be easily parallelized because it requires solving independent minimization problems.

\begin{algorithm}
\caption{Randomized PINN Algorithm}
\label{alg:randomized_pinn}
\begin{algorithmic}[1]
\Require{number of samples $N_{\text{ens}}$}
\For{$k = 1, \dots, N_{\text{ens}}$}
    \State Sample random noises $\boldsymbol\omega_r \sim \mathcal{N}(0, \sigma^2_r\mathbf{I})$, $\boldsymbol\omega_b \sim \mathcal{N}(0, \sigma^2_b\mathbf{I})$, $\boldsymbol\omega_y \sim \mathcal{N}(0, \sigma^2_y\mathbf{I})$, $\boldsymbol\omega_u \sim \mathcal{N}(0, \sigma^2_u\mathbf{I})$, $\boldsymbol\omega_\theta \sim \mathcal{N}(0, \sigma^2_p\mathbf{I})$, $\boldsymbol\omega_\phi\sim \mathcal{N}(0, \sigma^2_p\mathbf{I})$.
    \State Propose $\boldsymbol\theta_k$ by optimizing the randomized loss \eqref{eq:rPINN_loss}.
    \State Accept the sample and move to the next iteration.
\EndFor
\end{algorithmic}
\end{algorithm}

We note that the samples of the randomized minimization problem solution converge to the exact Bayesian posterior when the forward model is linear and the prior and likelihood distributions are Gaussian. 
However, the PINN model is non-linear in DNN parameters, and rPINN samples may deviate from the true posterior. The resulting bias can be removed using the Metropolis-Hastings (MH) algorithm~\cite{wang2018randomized,zong2023randomized}. In the MH algorithm, the samples are accepted or rejected according to the acceptance ratio, which can be derived in the closed form or approximated using the Jacobian of the map between the space of injected random noises and the space of model parameters and residuals. However, for high-dimensional rPINN problems, evaluating the acceptance ratio is computationally expensive. 
In other randomized-based methods (e.g., randomized MAP and randomized PICKLE), the rejection rates were very low \cite{wang2018randomized,zong2023randomized}. In this work, we also find that accepting all rPINN samples does not introduce significant bias by comparing the rPINN results with those obtained from other methods.  

To ensure a fair comparison,  we first select a set of weights in the deterministic PINN loss so that accurate MAP estimates with regard to the reference solutions and parameters can be obtained. These weights define the distributions of the noise terms in the rPINN loss function according to Eq \eqref{eq:rpinn_weights}. 


After samples of the posterior distribution are obtained, we can evaluate the ``predictive posterior'' probability distribution of the PINN solution. For example, the predictive posterior of the PDE state $u$ at $\mathbf{x}^*$ can be found as
\begin{eqnarray}\label{eq:BMA}
    P(u| \mathbf{x}^*, \mathcal{D}) = \int P(u|\mathbf{x}^*, \boldsymbol\theta)P(\boldsymbol\theta| \mathcal{D}) \, d\boldsymbol\theta = \mathbb{E}_{\boldsymbol\theta \sim P(\boldsymbol\theta | \mathcal{D})}[P(u|\mathbf{x}^*,  \boldsymbol\theta)] 
\end{eqnarray}
where $P(u|\mathbf{x}^*, \boldsymbol\theta)$ is the predictive distribution. In this work, we set $P(u|\mathbf{x}^*, \boldsymbol\theta)$ to be a Dirac delta function centered at the PINN prediction $\hat{u}(\mathbf{x}^*; \boldsymbol\theta)$. 
Then, the expectation and variance of the predictive posterior of $u$ can be estimated using the  rPINN samples 
$\{\boldsymbol\theta_i\}_{i=1}^{N_{\text{ens}}}$ as
\begin{eqnarray}\label{eq:BMA_mean}
   \mu_u (\mathbf{x}^* |  \mathcal{D})  
   \approx \frac{1}{N_{\text{ens}}} \sum_{k=1}^{N_{\text{ens}}} \hat{u}(\mathbf{x}^*; \boldsymbol\theta_k)
\end{eqnarray}
and 
\begin{eqnarray}\label{eq:BMA_var}
  \sigma_u^2( \mathbf{x}^* |  \mathcal{D})  
    \approx \frac{1}{N_{\text{ens}}-1} \sum_{k=1}^{N_{\text{ens}}} [\hat{u}(\mathbf{x}^*; \boldsymbol\theta_k) - \mu_u (\mathbf{x}^* |  \mathcal{D})  ]^2,
\end{eqnarray}
respectively. 
We use the relative $\ell^2$ and $\ell^\infty$ errors to assess the closeness of the approximate posterior mean predictions with respect to the reference solutions. For example, the relative $\ell^2$ and $\ell^\infty$ errors in the $u$ prediction are defined, respectively, as  
\begin{eqnarray}\label{eq:lp_error}
    r\ell^2_u  &=& \sqrt{\frac{\sum_{i=1}^{N}[\mu_u ( \mathbf{x}_i | \mathcal{D}) - u(\mathbf{x}_i)]^2}{\sum_{i=1}^{N}[u(\mathbf{x}_i)]^2}}, \\
    \ell^\infty_u  &=& \max_i \left( |\mu_u ( \mathbf{x}_i | \mathcal{D}) - u(\mathbf{x}_i)| \right) \quad i = 1, \dots, N,
\end{eqnarray}
where $N$ is the number of points within the domain where the reference solution is available. 

\section{Numerical Results}\label{sec:results}
In this section, we use the rPINN method for quantifying uncertainty in the PINN inverse solutions of linear and non-linear Poisson equations and the diffusion equation with space-dependent diffusion coefficient. We compare rPINN estimates of posterior distributions with those obtained with HMC and SVGD methods.  Additionally, we investigate the deep ensemble method for PINN training, where the \emph{deterministic} PINN loss function is minimized $N_{\text{ens}}$ times for different random initialization. We denote these methods as BPINN-HMC, BPINN-SVGD, and PINN-DE, respectively. 
In the Appendix, we provide details of the BPINN-HMC and BPINN-SVGD methods and the analysis of HMC convergence for different cases. 

\subsection{One-Dimensional Linear Poisson Equation}\label{sec:linear_poisson}
First, we consider the one-dimensional linear Poisson equation
\begin{align}\label{eq:1d_linear_poisson}
    k\frac{\partial^2 u}{\partial x^2} &= f(x), \quad x \in [-1, 1] \\
    u(x) &= u_l, \quad x = -1\\
    u(x) &= u_r, \quad x = 1.
\end{align}
We assume that the coefficient $k$ is known (we set to $k = -1/\pi^2$) and 
that the source function $f(x)$ and the boundary conditions are unknown. We also assume that $N_f$ noisy measurements of $f$ and the noisy estimates of each boundary condition are available.  
To generate the noisy measurements $\{ \tilde{f}(x^i_f) \}_{i = 1}^{N_f}$ and $\{ \tilde{u}(x^i_b) \}_{i = 1}^{N_b = 2} = \{ \tilde{u}_l, \tilde{u}_r \}$ and evaluate the accuracy of the considered methods, we assume the following reference forms of the source function and the boundary conditions: $f(x) = \sin(\pi x)$, $u_l = \sin(-\pi) $, and $u_r = \sin(\pi)$. The measurements are generated by adding independent random noises with variance $\sigma^2$ to the reference values of the respective variables.  For the reference $f(x)$, $u_l$, and $u_r$, Eq \eqref{eq:1d_linear_poisson} allows the analytical (reference) solution $u(x) = \sin(\pi x)$.  

In the examples below, we obtain the posterior distributions of $u(x)$ and $f(x)$ for $N_f=32$ and $128$ and two measurement noise variances $\sigma^2=10^{-4}$ and $10^{-2}$.  We approximate $u(x)$ and $f(x)$ with a DNN and its second derivative, respectively, as $u(x) \approx \hat{u}(x;\boldsymbol\theta)$ and $f(x) \approx \hat{f}(x,\boldsymbol{\theta}) = \partial^2 \hat{u}(x;\boldsymbol\theta)/ \partial x^2$. Here, the DNN has two hidden layers and 50 neurons per layer. The loss functions in the PINN and randomized PINN methods are, respectively,
\begin{align}\label{eq:det_pinn_loss_1d_linear_poisson}
 \boldsymbol\theta^* = \arg\min_\theta \Big\{ \frac{\lambda_f}{N_f} \sum_{i = 1}^{N_{f}}  \left [ \hat{f}(x^i_f; \boldsymbol\theta) - \tilde{f}(x^i_f)  \right ]^2 + \frac{\lambda_b}{N_b} \sum_{i = 1}^{N_{b}} [ \hat{u}(x^i_b; \boldsymbol\theta) - \tilde{u}(x^i_b)  ]^2  + \sum_{i=1}^{N_\theta} \theta_i^2     \Big\} ,
\end{align} 
and
\begin{eqnarray}\label{eq:pinn_loss_1d_linear_poisson}
 \boldsymbol\theta^* &=& \arg\min_\theta \Big\{ \frac{1}{\sigma^2_f} \sum_{i = 1}^{N_{f}}  \left [ \hat{f}(x^i_f; \boldsymbol\theta)  - \omega^i_f \right ]^2  \nonumber \\ 
&+& \frac{1}{\sigma^2_b} \sum_{i = 1}^{N_{b}} [ \hat{u}(x^i_b; \boldsymbol\theta) - \tilde{u}(x^i_b) -  \omega^i_b ]^2  + \frac{1}{\sigma^2_p} \sum_{i=1}^{N_\theta} (\theta_i^2 - \omega^i_p )   \Big\} .
\end{eqnarray} 
In the PINN loss function \eqref{eq:det_pinn_loss_1d_linear_poisson}, we find that $\lambda_f = 27000$ and $\lambda_b = 2700$ provide the smallest error in the PINN solution $\hat{u}$ with respect to the reference $u$. In the Poisson equation, the unknown $f(x)$ is additive, and Eq \eqref{eq:rpinn_weights} is replaced with:
\begin{eqnarray}
    \sigma^2_f = \sigma^2 , \quad
    \sigma^2_b = \sigma^2 \frac{\lambda_f N_b}{\lambda_b N_f},   \quad
    \sigma^2_p = \sigma^2 \frac{\lambda_f}{N_f}.
\end{eqnarray}
The resulting values of $\sigma_f$, $\sigma_b$, and $\sigma_p$ for each combination of $N_f$ and $\sigma$ are listed in Table~\ref{tab:1d_linear_poisson}. 
In rPINN and PINN-DE, the Adam stochastic optimization algorithm is used with a learning rate of \num{1e-3}. The details of HMC and SVGD simulations are given in \ref{sec:HMC} and \ref{sec:SVGD}, respectively.

\begin{figure}[!h]
    \includegraphics[width=0.95\textwidth]{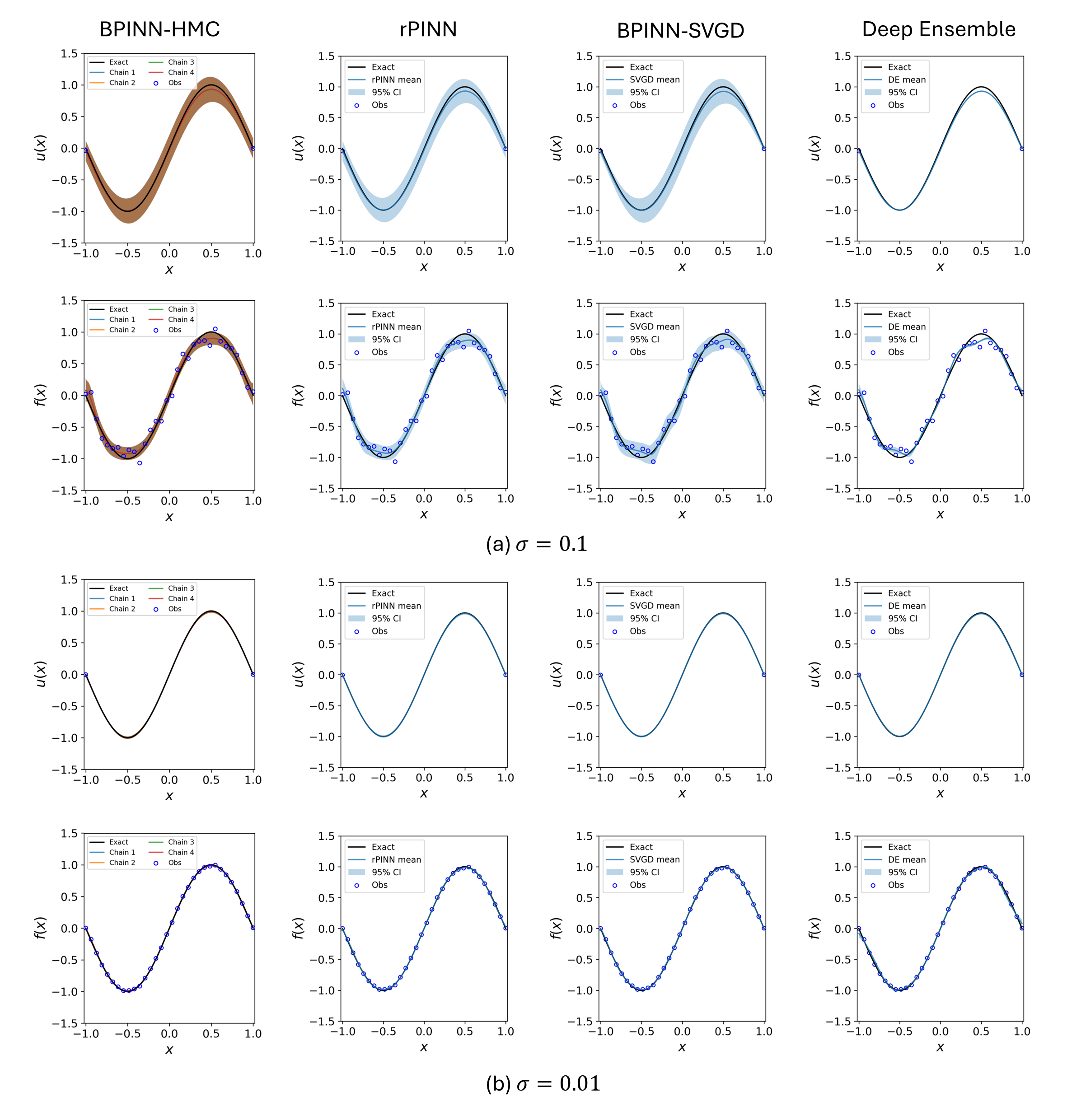}
    \caption{One-dimensional linear Poisson equation: posterior mean and confidence intervals of $u(x)$ and $f(x)$ computed from HMC (first column), rPINN (second column), SVGD (third column), and DE (fourth column) for (a) $\sigma = 0.1$ and, (b) $\sigma = 0.01$. The number of residual points is $N_f = 32$. The shaded areas represent $95\%$ confidence intervals (the pointwise mean plus/minus two standard deviations). The HMC chains produce fully overlapping confidence intervals. Therefore, the results for one chain can be seen on the HMC plots. }
    \label{fig:1d_linear_poisson}
\end{figure}

Figure~\ref{fig:1d_linear_poisson} shows mean estimates $\mu_u$ and $\mu_f$ and the confidence intervals estimated from the rPINN, BPINN-HMC, BPINN-SVGD, and PINN-DE methods. Table~\ref{tab:1d_linear_poisson} summarizes the accuracy of these methods in terms of the $r\ell^2$ and  $\ell^\infty$ error,  the spatial average of the standard deviations, the LPPs, and the coverage (the percentage of the domain where the reference solution is within the $95 \%$ confidence interval of the predictive distribution). In addition, we report the total execution time for obtaining $N_{\text{ens}} = 5000$ posterior samples. 

\begin{table}[htb!]
    \centering
    \small
    \caption{1D linear Poisson equation: comparisons of $u$ and $f$ statistics predicted by rPINN, BPINN-HMC, BPINN-SVGD, and PINN-DE for two measurement noise standard deviations  $\sigma = 0.1$ and $0.01$, and  $N_f = 32$ and $128$. The values of $\sigma_f, \sigma_b$ and $\sigma_p$ are also listed.}
    \begin{adjustbox}{max width=\textwidth, max height=\textheight}
    \begin{tabular}{p{1.2cm} l l S[table-format=1.2e-1] S[table-format=1.2e-1] S[table-format=1.2e-1] S[table-format=5] S[table-format=3] l}
        \toprule
          & & \quad \textbf{Method} &\textbf{$r\ell^2$ error} & \textbf{$\ell^\infty$ error} & \textbf{Ave. Std} & \textbf{LPP} & \textbf{Coverage} & \textbf{Time} \\
        \midrule
        \multirow{8}{*}{$N_f = 32$} & \multicolumn{8}{l}{\textbf{Measurement Noise Std:} $\sigma = 0.1 \rightarrow \sigma_f = 0.1, \; \sigma_b = 0.079, \; \sigma_p = 2.905$} \\
        \cmidrule(lr){2-9}
        & \multirow{4}{*}{$u(x)$} & \quad rPINN & 5.1e-2 & 7.0e-2 & 9.7e-2 & 742 & 100\% & 416s \\
        & &\quad BPINN-HMC & 5.1e-2 & 7.0e-2 & 9.8e-2 & 740 & 100\% & 4036s \\
        & &\quad BPINN-SVGD & 5.6e-2 & 7.4e-2 & 9.2e-2 & 785 & 100\% & 1801s \\
        & &\quad PINN-DE & 5.3e-2 & 7.2e-2 & 4.7e-3 & -3580 & 20\% & 418s \\
        \cmidrule(lr){2-9}
        & \multirow{4}{*}{$f(x)$} & \quad rPINN & 7.3e-2 & 1.2e-1 & 5.8e-2 & 881 & 97\% & \\
        & & \quad BPINN-HMC & 7.5e-2 & 1.4e-1 & 5.4e-2 & 894 & 94\% & \\
        & & \quad BPINN-SVGD & 8.5e-2 & 1.2e-1 & 6.9e-2 & 811 & 99\% & \\
        & & \quad PINN-DE & 8.2e-2 & 1.3e-1 & 1.5e-2 & -798 & 48\% & \\
        \midrule
        \multirow{8}{*}{$N_f = 32$} & \multicolumn{8}{l}{\textbf{Measurement Noise Std:} $\sigma = 0.01 \rightarrow  \sigma_f = 0.01, \;  \sigma_b = 0.008, \;  \sigma_p = 0.291$} \\
        \cmidrule(lr){2-9}
        & \multirow{4}{*}{$u(x)$} & \quad rPINN & 5.7e-3 & 7.8e-3 & 9.8e-3 & 1661 & 100\% & 428s \\
        & & \quad BPINN-HMC & 5.6e-3 & 7.7e-3 & 9.6e-3 & 1666 & 100\% & 12365s \\
        & & \quad BPINN-SVGD & 5.1e-3 & 6.9e-3 & 8.8e-3 & 1704 & 100\% & 4320s \\
        & & \quad PINN-DE & 5.5e-3 & 7.7e-3 & 1.6e-3 & 1725 & 53\% & 427s \\
        \cmidrule(lr){2-9}
        & \multirow{4}{*}{$f(x)$} & \quad rPINN & 9.6e-3 & 1.6e-2 & 5.8e-3 & 1746 & 90\% & \\
        & & \quad BPINN-HMC & 1.0e-2 & 1.5e-2 & 5.4e-3 & 1733 & 84\% & \\
        & & \quad BPINN-SVGD & 1.1e-2 & 2.1e-2 & 5.4e-3 & 1718 & 85\% & \\
        & & \quad PINN-DE & 1.3e-2 & 3.1e-2 & 4.0e-3 & 1377 & 61\% & \\
        \midrule
        \multirow{8}{*}{$N_f = 128$} & \multicolumn{8}{l}{\textbf{Measurement Noise Std:} $\sigma = 0.1 \rightarrow  \sigma_f = 0.1, \;  \sigma_b = 0.04, \;  \sigma_p = 1.452$} \\
        \cmidrule(lr){2-9}
        & \multirow{4}{*}{$u(x)$} & \quad rPINN & 2.5e-2 & 4.9e-2 & 4.9e-2 & 1015 & 100\% & 1038s \\
        & & \quad BPINN-HMC & 2.7e-2 & 5.0e-2 & 4.8e-2 & 1016 & 100\% & 34533s \\
        & & \quad PINN-SVGD & 3.0e-2 & 4.9e-2 & 4.5e-2 & 1043 & 96\% & 9642s \\
        & & \quad PINN-DE & 3.2e-2 & 5.0e-2 & 4.4e-3 & -1475 & 11\% & 1028s \\
        \cmidrule(lr){2-9}
        & \multirow{4}{*}{$f(x)$} & \quad rPINN & 4.7e-2 & 8.3e-2 & 4.0e-2 & 1042 & 98\% & \\
        & & \quad BPINN-HMC & 3.8e-2 & 8.6e-2 & 2.8e-2 & 1180 & 97\% & \\
        & & \quad BPINN-SVGD & 4.0e-2 & 8.6e-2 & 4.7e-2 & 1002 & 100\% & \\
        & & \quad PINN-DE & 3.5e-2 & 1.1e-1 & 1.5e-2 & 1252 & 71\% & \\
        \midrule
        \multirow{8}{*}{$N_f = 128$} & \multicolumn{8}{l}{\textbf{Measurement Noise Std:} $\sigma = 0.01 \rightarrow  \sigma_f = 0.01, \;  \sigma_b = 0.004, \;  \sigma_p = 0.145$} \\
        \cmidrule(lr){2-9}
        & \multirow{4}{*}{$u(x)$} & \quad rPINN & 2.8e-3 & 4.7e-3 & 5.2e-3 & 1917 & 100\% & 1041s \\
        & & \quad BPINN-HMC & 2.9e-3 & 4.8e-3 & 5.8e-3 & 1872 & 100\% & 41715s \\
        & & \quad BPINN-SVGD & 3.0e-3 & 4.9e-3 & 5.1e-3 & 1929 & 100\% & 9832s \\
        & & \quad PINN-DE & 4.1e-3 & 5.1e-3 & 1.7e-3 & 2085 & 63\% & 994s \\
        \cmidrule(lr){2-9}
        & \multirow{4}{*}{$f(x)$} & \quad rPINN & 1.0e-2 & 3.5e-2 & 4.5e-3 & 1841 & 87\% & \\
        & & \quad  BPINN-HMC & 4.7e-3 & 1.1e-2 & 2.7e-3 & 2039 & 92\% & \\
        & & \quad BPINN-SVGD & 4.2e-3 & 7.3e-3 & 3.8e-3 & 1994 & 100\% & \\
        & & \quad PINN-DE & 2.1e-2 & 7.0e-2 & 5.6e-3 & 1584 & 70\% & \\
        \bottomrule
    \end{tabular}
    \end{adjustbox}
    \label{tab:1d_linear_poisson}
\end{table}

Results in Table~\ref{tab:1d_linear_poisson} and Figure~\ref{fig:1d_linear_poisson} show that for the considered values of $\sigma$ and $N_f$, the four methods yield similar estimates of $\mu_u$ and $\mu_f$. The estimates of average $\sigma_u$ and $\sigma_f$ are similar in the rPINN, BPINN-HMC, and BPINN-SVGD methods and are an order of magnitude smaller in the PINN-DE method. For the larger measurement noise, LPP in the PINN-DE method is seven orders of magnitude smaller than LPPs in the other three methods, i.e., PINN-DE significantly underestimates uncertainty. For the smaller measurement noise, LPPs in all four methods are of the same order.  
This is because the PINN-DE method does not account for the measurement uncertainty and only partially accounts for the PINN model uncertainty (i.e., model uncertainty due to PINN model initialization). The relative importance of measurement uncertainty with respect to uncertainty due to initialization decreases as measurement noise diminishes.


Even though HMC produces similar means, variances, and LPPs of $u$ and $f$ compared to rPINN and SVGD for all considered combinations of $N_f$ and $\sigma$, the analysis of HMC statistics shows the lack of convergence for the case with $N_f = 128$ and $\sigma = 0.01$. This indicates that even for linear PDE problems, HMC must be applied very carefully. The details of the HMC simulations and the convergence analysis are given in \ref{sec:HMC}. 

Table \ref{tab:1d_linear_poisson} also indicates that the HMC and SVGD execution times strongly depend on $N_f$ and $\sigma$.  On the other hand, the rPINN execution time is practically independent of $\sigma$ and has a weaker dependence on $N_f$ than the other two methods. The former is due to the Monte Carlo nature of rPINN, and the latter is because of the linear dependence of the cost of the rPINN loss function computations on $N_f$. For this problem, on average, the rPINN execution time is approximately 27 times less than HMC and eight times smaller than SVGD.

As expected, we find that smaller $\sigma$ (i.e., more reliable observations) yields more informative (higher LPP) and less uncertain (smaller variance) posterior distributions of $u$ and $f$. Furthermore, we observe that as $N_f$ increases from $32$ to $128$, the relative error and the average standard deviation of rPINN, BPINN-HMC, and BPINN-SVGD $u(x)$ predictions are nearly halved. Additionally, LPP increases by  $\approx 35\%$ for $\sigma = 0.1$ and 15\% for $\sigma = 0.01$. However, the average standard deviation in the $f$ solution decreases only slightly. 
We note that if the likelihood parameters are chosen according to Eq.~\eqref{eq:likelihood_variances_uniform},  the $r\ell^2_u$ and $r\ell^2_f$ errors increase and LPPs decrease with increasing $N_f$.

\subsection{One-Dimensional Non-Linear Poisson Equation}\label{sec:nonlinear_poisson}
Here, we consider the one-dimensional non-linear Poisson equation:
\begin{align}\label{eq:1d_nonlinear_poisson}
        \lambda \frac{\partial^2 u}{\partial x^2} + k \tanh (u) &= f(x) \quad x \in [-0.7, 0.7], \\
    u(x) &= u_l, \quad x = -0.7\\
    u(x) &= u_r, \quad x = 0.7
\end{align}
where $\lambda$ and $k$ are assumed to be known and set to $\lambda = 0.01$ and $k = 0.7$. Furthermore, we assume that the source function and boundary conditions are unknown and their noisy measurements are available. We set the reference forms of boundary conditions and the source function to $u_l =\sin(-4.2)^3$,  $u_r = \sin(4.2)^3$, and
\begin{equation*}
  f(x) = \lambda[-108\sin(6x)^3 + 216\sin(6x)\cos(6x)^2] + k \tanh(u(x)).
\end{equation*}
The corresponding reference $u$ solution is $u(x)=\sin(6x)^3$. As in the previous example, the reference values are used to generate $\tilde{u}_l$, $\tilde{u}_r$, and equally spaced $f$ measurements $\{ \tilde{f}(x^i_f) \}_{i = 1}^{N_f}$ with added independent random noise with the standard deviation $\sigma$.  

We set $N_f = 32$ and consider $\sigma = 0.01$ and $0.1$. We use the DNN $\hat{u}(x; \boldsymbol\theta)$ with two hidden layers and 50 neurons per layer to represent $u(x)$. The source function is approximated as $f(x) \approx \hat{f}(x; \boldsymbol\theta) = 
\lambda \frac{\partial^2 \hat{u}(x; \boldsymbol\theta)}{\partial x^2} + k\tanh{(\hat{u}(x; \boldsymbol\theta))} 
$. Then, the PINN and rPINN loss functions for the non-linear Poisson equation are given by Eqs \eqref{eq:det_pinn_loss_1d_linear_poisson} and \eqref{eq:pinn_loss_1d_linear_poisson}, respectively.
We use the same weights $\lambda_f = 27000$, and $\lambda_b = 2700$ as in the PINN solution of the linear Poisson equation, as they provide a near-optimal MAP estimate of $u$ in terms of $r\ell^2_u$. The resulting values of $\sigma_f $, $\sigma_b$, $\sigma_p$ are listed in Table \ref{tab:1d_nonlinear_poisson}. 

\begin{table}[htb!]
    \centering
    \small
    \caption{One-dimensional nonlinear Poisson equation: comparisons of $u$ and $f$ statistics predicted by rPINN, BPINN-SVGD, and PINN-DE for  $\sigma^2 = 10^{-4}$ and $10^{-2}$ and $N_f = 32$. The  values of $\sigma_f, \sigma_b$ and $\sigma_p$ are also listed.}
    \begin{adjustbox}{max width=\textwidth, max height=\textheight}
    \begin{tabular}{l l S[table-format=1.2e-1] S[table-format=1.2e-1] S[table-format=1.2e-1] S[table-format=5] S[table-format=3]}
        \toprule
        & \quad \textbf{Method} &\textbf{$r\ell^2$ error} & \textbf{$\ell^\infty$ error} & \textbf{Ave. Std} & \textbf{LPP} & \textbf{Coverage}  \\
        \midrule
       \multicolumn{7}{l}{\textbf{Measurement Noise Std:} $\sigma = 0.1 \rightarrow \sigma_f = 0.1, \; \sigma_b = 0.079, \; \sigma_p = 2.905$} \\
        \hline
        \multirow{3}{*}{$u(x)$} & \quad rPINN & 1.9e-1 & 1.7e-1 & 2.5e-1 & 367 & 100\%  \\
        &\quad BPINN-SVGD & 2.6e-0 & 1.9e-0 & 3.1e-0 & -606 & 100\%  \\
        &\quad PINN-DE & 1.8e-1 & 2.1e-1 & 1.5e-1 & 569 & 100\%  \\
        \hline
        \multirow{3}{*}{$f(x)$} & \quad rPINN & 1.9e-1 & 2.4e-1 & 9.0e-2 & 653 & 96\%  \\
        & \quad BPINN-SVGD & 1.5e-1 & 1.6e-1 & 1.0e-1 & 668 & 98\%  \\
        & \quad PINN-DE & 1.8e-1 & 2.1e-1 & 4.0e-2 & 397 & 58\%  \\
        \midrule
         \multicolumn{7}{l}{\textbf{Measurement Noise Std:} $\sigma = 0.01 \rightarrow  \sigma_f = 0.01, \;  \sigma_b = 0.008, \;  \sigma_p = 0.291$} \\
        \hline
         \multirow{3}{*}{$u(x)$} & \quad rPINN & 1.4e-2 & 1.4e-2 & 8.7e-2 & 861 & 100\%   \\
        & \quad BPINN-SVGD & 2.7e-0 & 2.1e-0 & 3.1e-0 & -594 & 100\%  \\
        & \quad PINN-DE & 3.1e-2 & 3.9e-2 & 9.2e-2 & 796 & 100\%   \\
        \hline
         \multirow{3}{*}{$f(x)$} & \quad rPINN & 2.6e-2 & 3.1e-2 & 9.5e-3 & 1487 & 86\% \\
        & \quad BPINN-SVGD & 1.9e-2 & 2.2e-2 & 1.4e-2 & 1481 & 100\%  \\
        & \quad PINN-DE & 2.9e-2 & 3.8e-2 & 1.5e-2 & 1318 & 100\%  \\
        \bottomrule
    \end{tabular}
    \end{adjustbox}
    \label{tab:1d_nonlinear_poisson}
\end{table}

\begin{figure}[!htb]
	\includegraphics[width=0.95\textwidth]{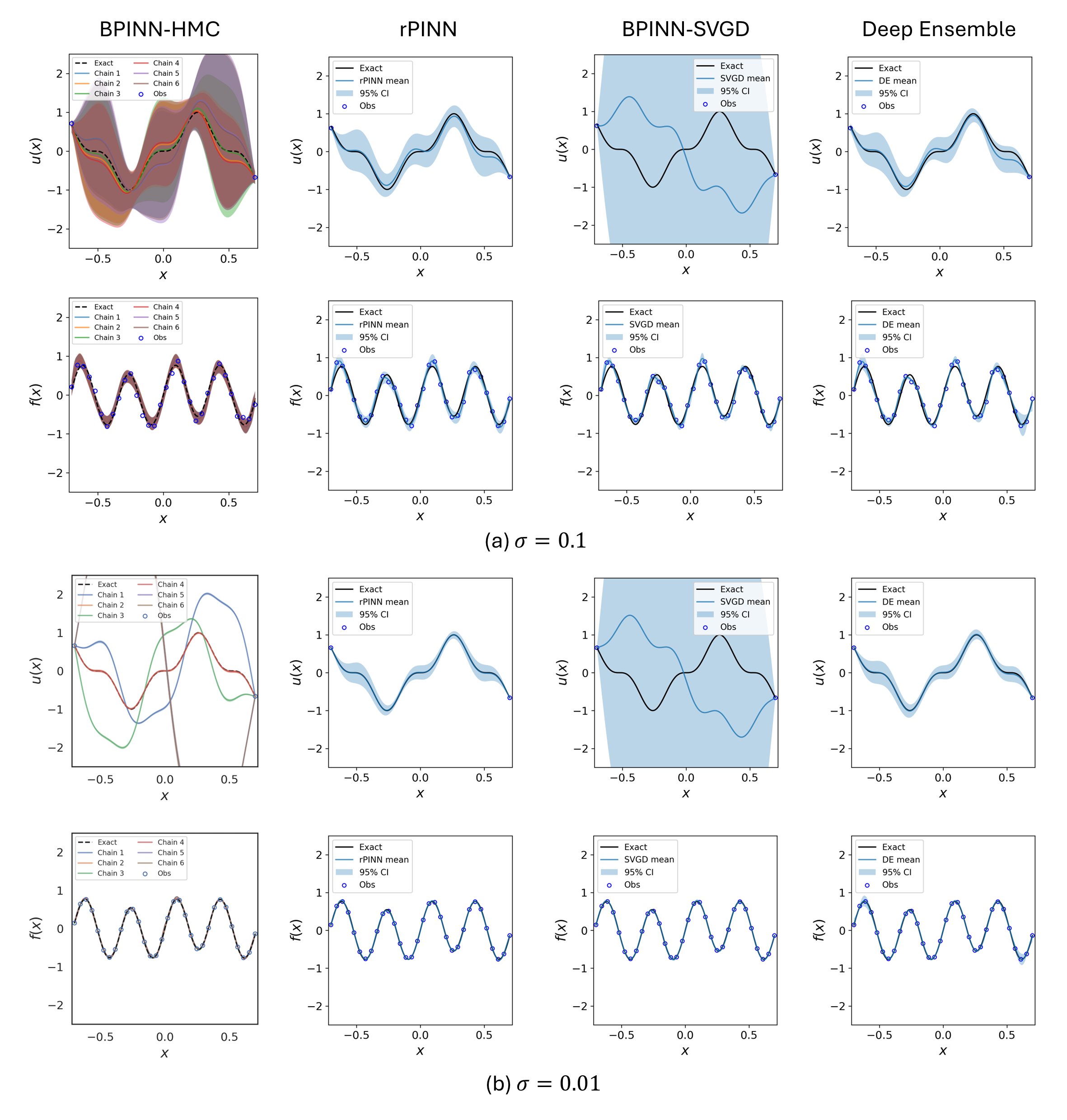}
    \caption{ One-dimensional non-linear Poisson problem: posterior predictions of $u(x)$ and $f(x)$ given by HMC (first column), rPINN (second column), SVGD (third column), and DE (fourth column) for (a) $\sigma = 0.1$ and (b) $\sigma = 0.01$. The number of residual points is $N_f = 32$.}
    \label{fig:1d_nonlinear_poisson}
\end{figure}

Figure~\ref{fig:1d_nonlinear_poisson} shows the estimated mean of $u$ and $f$ and the associated confidence intervals computed from the four methods. Table~\ref{tab:1d_nonlinear_poisson} summarizes the accuracy of rPINN, BPINN-SVGD, and PINN-DE in terms of the $r\ell^2$ and $\ell^\infty$ errors, the spatial average of the standard deviations, the LPPs, and the coverage. In the HMC study, we run six chains. We find that different chains yield different $u$ distributions. Therefore, we report $r\ell^2_u$, $r\ell^2_f$ and LPPs for each HMC chain in Table \ref{tab:1d_nonlinear_poisson_hmc}, and show mean and the confident intervals obtained from each chain in Figure \ref{fig:1d_nonlinear_poisson}.

The rPINN and the PINN-DE methods have small $r\ell^2_u$ and $r\ell^2_f$ consistent with the relative error in the PINN MAP predictions. The average standard deviations decrease and the LPPs of $u$ and $f$ increase with decreasing $\sigma$. For both values of $\sigma$, the reference $u$ and $f$ are mostly within the predicted confidence intervals. This indicates that rPINN yields informative posterior distributions of $u$ and $f$. It is interesting to see that $\sigma^2_u$ predicted by rPINN and PINN-DE are of the same order indicating that the initialization uncertainty dominates the PINN solution's total uncertainty.

On the other hand, we find that HMC chains fail to converge, and SVGD provides a non-informative posterior of the (unobserved) state $u$ and only yields the informative posterior of the observed source function $f$. In the HMC study for $\sigma = 0.1$, the difference between the maximum and minimum $r\ell^2_u$ found from the six chains is approximately 150 times larger than that of $f$, and the maximum difference in the LPP of $u$ between chains is about 40 times larger than that of $f$. For $\sigma = 0.01$, the six chains yield an even larger range of $r\ell^2_u$ and LPPs than those of $f$. The difference between the maximum and minimum $r\ell^2_u$ is approximately 350 times larger than that of $f$, and the maximum difference in the LPPs of $u$ between chains is about 1000 times larger than that of $f$. Specifically, three chains (chains 1, 3, and 6) produce the posterior distributions of $u$ that significantly deviate from each other and the reference $u$. The variability in the posterior $u$ distributions and consistency in the $f$ distributions are illustrated in Figure~\ref{fig:1d_nonlinear_poisson}.

The analysis of HMC results showing the lack of convergence is given in \ref{sec:HMC}. This analysis suggests that different HMC chains do not mix. They sample the multimodal posterior of the DNN parameters around different isolated modes, all corresponding to accurate $f$ predictions but $u$ predictions of variable quality.
We hypothesize that SVGD also fails because of the multimodality of the $u$ posterior distribution and the isolated nature of these modes. SVGD samples cluster around local modes, and the space between the modes in not sufficiently sampled. 

\begin{table}[!htb]
\centering
\caption{One-dimensional non-linear Poisson equation: summary of $u$ and $f$ statistics predicted by different HMC chains for $\sigma^2 = 10^{-4}$ and $10^{-2}$.}
{\small
\begin{tabular}{p{1.5cm}p{2cm}p{2cm}p{2cm}p{2cm}}
\hline
& $r\ell_u^2$ error & LPP $u$ & $r\ell_f^2$ error $f$ & LPP $f$ \\
\hline
 \multicolumn{5}{l}{Measurement Noise Std: $\sigma = 0.1$ } \\
\hline
chain 1  & \num{5.73e-1} &  \num{85} & \num{1.57e-1} &   \num{724} \\
chain 2  & \num{2.78e-1} &  \num{178} & \num{1.59e-1} &  \num{723} \\    
chain 3  & \num{1.17e-1} &  \num{109} & \num{1.58e-1} &  \num{724}  \\   
chain 4  & \num{3.58e-1} &  \num{170} & \num{1.59e-1} & \num{722}  \\   
chain 5  & \num{5.51e-1} &  \num{58} & \num{1.60e-1} &  \num{724} \\   
chain 6  & \num{5.28e-1} &  \num{76} & \num{1.58e-1} &  \num{725} \\   
\hline
 \multicolumn{5}{l}{Measurement Noise Std: $\sigma = 0.01$} \\
\hline
chain 1  & \num{1.77e-0} &  \num{-274551} & \num{2.76e-2} &  \num{1415} \\
chain 2  & \num{1.53e-2} &  \num{1483} & \num{2.85e-2} &  \num{1412} \\    
chain 3  & \num{1.76e-0} &  \num{-264069} & \num{2.82e-2} &  \num{1400}  \\   
chain 4  & \num{1.55e-2} &  \num{1484} & \num{2.86e-2} &  \num{1410}  \\   
chain 5  & \num{1.55e-2} &  \num{1483} & \num{2.85e-2} &   \num{1409} \\   
chain 6  & \num{8.23e-0} &  \num{-2924888} & \num{4.39e-2} &  \num{1126} \\       
\hline
\end{tabular}
}
\label{tab:1d_nonlinear_poisson_hmc}
\end{table}

\subsection{Two-Dimensional Diffusion Equation with Space-dependent Diffusion Coefficient}\label{sec:darcy}
Here, we consider the diffusion equation with space-dependent diffusion coefficient $k(\mathbf{x})$ 
defined on the rectangular domain $\Omega=[0,L_1]\times [0,L_2]$:
\begin{align}
\nabla \cdot \left[ k(\mathbf{x}) \nabla h(\mathbf{x}) \right] &=  0, \quad \mathbf{x} \in \Omega  \label{eq:gwf} \\
 h(\mathbf{x}) &= H, \quad {x_1} = L_{1} \label{eq:gwf_dbc} \\
 -k(\mathbf{x}) \partial h(\mathbf{x}) /  \partial x_1 & = q, \quad x_1 = 0 \label{eq:gwf_nbc1} \\
 -k(\mathbf{x}) \partial h(\mathbf{x}) /  \partial x_2 & = 0, \quad x_2 = 0 \:\& \: x_2 = L_2. \label{eq:gwf_nbc2} 
 \end{align}
where $H$ and $q$ are the unknown Dirichlet and Neumann boundaries, respectively, and $k(\mathbf{x})$ is an unknown coefficient. Among many other applications, this equation describes flow in porous media. 

We apply the rPINN approach to obtain the Bayesian estimates of $y(\mathbf{x})=\ln k(\mathbf{x})$ and $h(\mathbf{x})$ given $N_y$ and $N_h$ noisy measurements of $y$ and $h$, respectively. There are two main reasons for estimating $y$ rather than $k$: this guarantees that $k(\mathbf{x})$ is positive; and, with application to groundwater flow, multiple field observations confirm that $\ln k$ has a Gaussian distribution. In the Bayesian framework, we use this body of knowledge to assume that the prior distribution of $y$ is Gaussian, and the mean and covariance of $y$ are available (e.g., from the geostatistical analysis of field measurements). 

\begin{figure}[htb!]
	\centering
	\subfloat[] {\includegraphics[angle=0,width=0.4\textwidth]{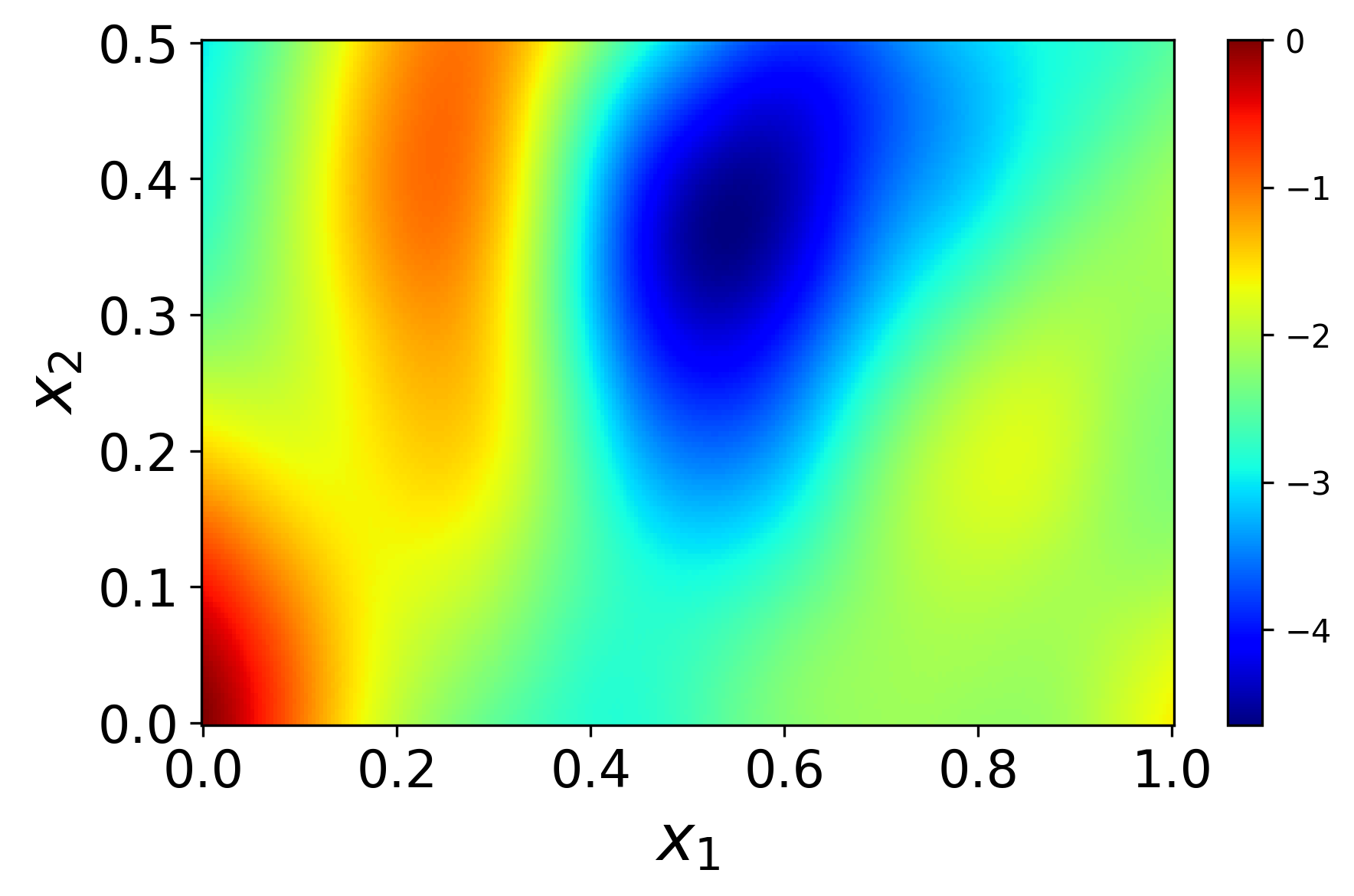}}
	\subfloat[] {\includegraphics[angle=0,width=0.4\textwidth]{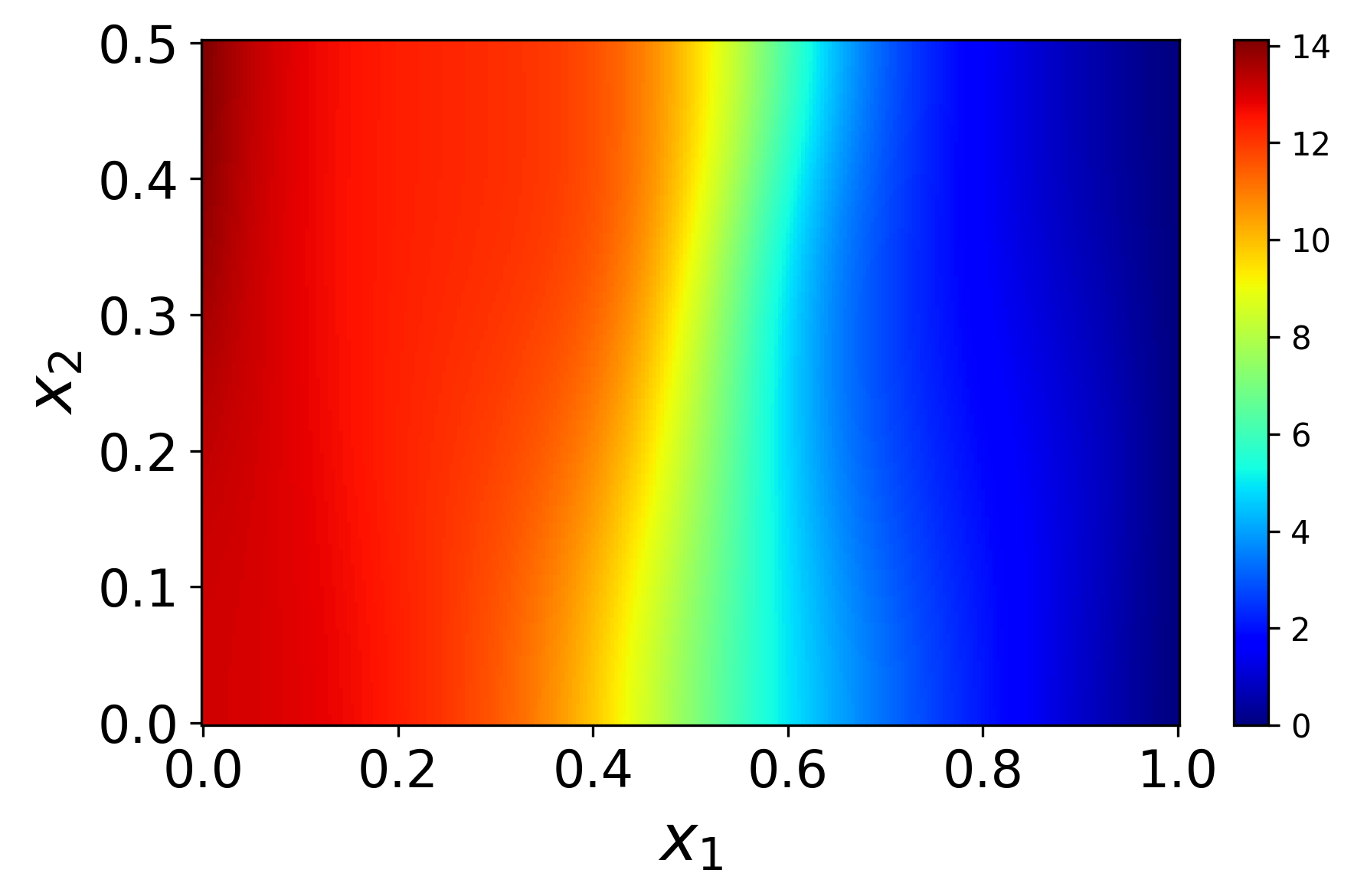}}
	\caption{Two-dimensional diffusion equation: (a) the reference $y_{\text{ref}}(\mathbf{x})=\ln k_{\text{ref}}(\mathbf{x}) $ field with a correlation length $\lambda=0.5$ and 
	(b) the reference hydraulic head field $h_{\text{ref}}(\mathbf{x})$.}
	\label{fig:reference_field}
\end{figure}

In this example, we assume that the (prior) mean and covariance of $y$ are $\hat{\mu}_y = -3$ and $\text{C}_y(x, x^{\prime}) = \hat{\sigma}^2_y \exp(-\frac{{\|x - x^{\prime}\|}^2}{\lambda^2 })$, correspondingly, with the variance $\hat{\sigma}^2_y = 0.81$ and the correlation length $\lambda = 0.5$. The domain size and the reference boundary conditions are set to $L_1 = 1$, $L_2 = 0.5$, $H=0$, and $q=1$. The reference field $y_{\text{ref}}$ is generated as a realization of the Gaussian field, and the reference hydraulic head field $h_{\text{ref}}$ is obtained by solving the diffusion equation \eqref{eq:gwf}-\eqref{eq:gwf_nbc2}, discretized with the finite-difference (FD) method as implemented in MODFLOW~\cite{harbaugh2005modflow}. The reference FD solution is obtained on a uniform $256\times128$ grid. The reference fields $y_{\text{ref}}$ and $h_{\text{ref}}$ are shown in Figure \ref{fig:reference_field}.

To generate noisy measurements, we sample $y_{\text{ref}}(\mathbf{x})$ and $h_{\text{ref}}(\mathbf{x})$ at $N_y$ and $N_h$ measurement locations and add independent zero-mean Gaussian noises with a standard deviation $\sigma$. Without loss of generality, we assumed that the number of and locations of $h$ and $y$ measurements are the same. We have $N_{dbr}$ uniformly spaced noisy head measurements at the right Dirichlet boundary and $N_{nbl}$ noisy measurements of the input flux at the left inhomogeneous Neumann boundary, with the same standard deviation $\sigma$. Here, we set $N_{nbl} = N_{dbr} =  54$. 
The PDE and boundary terms are evaluated at $N_r = 500$ and $N_{nbt} = N_{nbb} = 128$ residual points.  We use two separate DNNs with four hidden layers and 60 neurons per layer to represent $y(\mathbf{x})$ and $h(\mathbf{x})$:
\begin{align}
 y(\mathbf{x}) &\approx \hat{y}(\mathbf{x}; \boldsymbol\phi) := 
                     \mathcal{NN}_y(\mathbf{x}; \boldsymbol\phi) \\
    h(\mathbf{x}) &\approx \hat{h}(\mathbf{x}; \boldsymbol\theta) := 
                     \mathcal{NN}_h(\mathbf{x}; \boldsymbol\theta).
\end{align}
The PINN loss function takes the form:
\begin{align}\label{eq:pinn_loss_gwf}
(\boldsymbol\theta^*, \boldsymbol\phi^*) &=  \arg\min_{\boldsymbol\theta, \boldsymbol\phi}  \Big\{ \frac{\lambda_r}{N_r} \sum_{i = 1}^{N_r}  \mathcal{R}_{L}(\mathbf{x}^i_r; \boldsymbol\theta, \boldsymbol\phi)^2 + \frac{\lambda_{dbr}}{N_{dbr}} \sum_{i = 1}^{N_{dbr}} [\hat{h}(\mathbf{x}^i_{dbr}; \boldsymbol\theta) - \tilde{H}^i]^2 + \nonumber \\
&+ \frac{\lambda_{nbl}}{N_{nbl}} \sum_{i = 1}^{N_{nbl}} [\mathcal{R}_{nbl}(\mathbf{x}^i_{nbl}; \boldsymbol\theta, \boldsymbol\phi) - \tilde{q}^i ]^2 + \frac{\lambda_{nbt}}{N_{nbt}} \sum_{i = 1}^{N_{nbt}} \mathcal{R}_{nbt}(\mathbf{x}^i_{nbt}; \boldsymbol\theta, \boldsymbol\phi)^2 \nonumber \\
&+ \frac{\lambda_{nbb}}{N_{nbb}} \sum_{i = 1}^{N_{nbb}} \mathcal{R}_{nbb}(\mathbf{x}^i_{nbb}; \boldsymbol\theta, \boldsymbol\phi)^2 + \frac{\lambda_y}{N_y} \sum_{i = 1}^{N_y} [\hat{y}(\mathbf{x}^i_{y}; \boldsymbol\phi) - \tilde{y}(\mathbf{x}^i_{y})]^2 \nonumber \\
&+  \frac{\lambda_h}{N_h} \sum_{i = 1}^{N_h} [\hat{h}(\mathbf{x}^i_{h}; \boldsymbol\theta) - \tilde{h}(\mathbf{x}^i_{h})]^2 + \sum_{i = 1}^{N_\theta} \theta_i^2 + \sum_{i = 1}^{N_\phi} \phi_i^2\Big\} , 
\end{align} 
where $\lambda_{k}$ ($k = r, dbr, nbl, nbt, nbb, h, y$) refers to the PINN weights of the terms for the PDE residual, right Dirichlet boundary, left Neumann boundary, top Neumann boundary, bottom Neumann boundary, $h$ measurements, $y$ measurements respectively, $N_{NN}$ is the number of parameters in the DNNs, and the residuals are defined as
\begin{align}
\mathcal{R}_{L}(\mathbf{x}^i_r; \boldsymbol\theta, \boldsymbol\phi) &= e^{\hat{y}(\mathbf{x}^i_r; \boldsymbol\phi)} \nabla \hat{h}(\mathbf{x}; \boldsymbol\theta)\Big\vert_{\mathbf{x} = \mathbf{x}^i_r}, \nonumber \\
\mathcal{R}_{nbl}(\mathbf{x}^i_{nbl}; \boldsymbol\theta, \boldsymbol\phi) &= -e^{\hat{y}(\mathbf{x}^i_{nbl};\boldsymbol\phi)} \frac{\partial h(\mathbf{x}^i_{nbl}; \boldsymbol\theta)}{\partial x_1} \Big\vert_{x_1 = 0}, \nonumber \\
\mathcal{R}_{nbt}(\mathbf{x}^i_{nbt}; \boldsymbol\theta, \boldsymbol\phi) &= -e^{y(\mathbf{x}^i_{nbt};\boldsymbol\phi)} \frac{\partial h(\mathbf{x}^i_{nbt}; \boldsymbol\theta)}{\partial x_2} \Big\vert_{x_2 = L_2}, \nonumber \\
\mathcal{R}_{nbb}(\mathbf{x}^i_{nbb}; \boldsymbol\theta, \boldsymbol\phi) &= -e^{y(\mathbf{x}^i_{nbb};\boldsymbol\phi)} \frac{\partial h(\mathbf{x}^i_{nbb}; \boldsymbol\theta)}{\partial x_2} \Big\vert_{x_2 = 0}. \nonumber
\end{align}
The rPINN loss function is,
\begin{align}\label{eq:rpinn_loss_gwf}
(\boldsymbol\theta^*, \boldsymbol\phi^*) &=  \arg\min_{\boldsymbol\theta, \boldsymbol\phi}  \Big\{ \frac{1}{\sigma_r^2} \sum_{i = 1}^{N_r} [ \mathcal{R}_{L}(\mathbf{x}^i_r; \boldsymbol\theta, \boldsymbol\phi) - \omega^i_r]^2 + \frac{1}{\sigma_{dbr}^2} \sum_{i = 1}^{N_{dbr}} [\hat{h}(\mathbf{x}^i_{dbr}; \boldsymbol\theta) - \tilde{H}^i -\omega^i_{dbr}]^2 + \nonumber \\
&+ \frac{1}{\sigma_{nbl}^2} \sum_{i = 1}^{N_{nbl}} [\mathcal{R}_{nbl}(\mathbf{x}^i_{nbl}; \boldsymbol\theta, \boldsymbol\phi) - \tilde{q}^i -\omega^i_{nbl} ]^2 + \frac{1}{\sigma_{nbt}^2} \sum_{i = 1}^{N_{nbt}} [\mathcal{R}_{nbt}(\mathbf{x}^i_{nbt}; \boldsymbol\theta, \boldsymbol\phi) -  \omega^i_{nbt}]^2 \nonumber \\
&+ \frac{1}{\sigma_{nbb}^2} \sum_{i = 1}^{N_{nbb}} [\mathcal{R}_{nbb}(\mathbf{x}^i_{nbb}; \boldsymbol\theta, \boldsymbol\phi) - \omega^i_{nbb} ]^2 + \frac{1}{\sigma_{y}^2} \sum_{i = 1}^{N_y} [\hat{y}(\mathbf{x}^i_{y}; \boldsymbol\phi) - \tilde{y}(\mathbf{x}^i_{y}) - \omega^i_y]^2 \nonumber \\
&+  \frac{1}{\sigma_{h}^2} \sum_{i = 1}^{N_h} [\hat{h}(\mathbf{x}^i_{h}; \boldsymbol\theta) - \tilde{h}(\mathbf{x}^i_{h}) -  \omega^i_h]^2 +  \frac{1}{\sigma_{p}^2}\sum_{i = 1}^{N_{\theta}} [\theta_i - \omega^i_p] ^2 +  \frac{1}{\sigma_{p}^2}\sum_{i = 1}^{N_{\phi}} [\phi_i - \omega^i_p]^2\Big\} , 
\end{align} 
where $\omega_{k}$ ($k = r, dbr, nbl, nbt, nbb, h, y, p$) are the random noise variables. We empirically select weights in the deterministic PINN loss function as  $\lambda_r = \lambda_{dbr} = \lambda_{nbl} = \lambda_{nbt} = \lambda_{nbb} = \lambda_{h} = \lambda_{y} = N_{\theta} = N_{\phi}$. Then, we compute $\sigma^2_p$ from Eq \eqref{eq:sigmap} and the rest of the weights in the rPINN loss function from Eq \eqref{eq:sigmap} as
\eqref{eq:rpinn_weights} as 
\begin{eqnarray}
    \sigma^2_r = \sigma^2 \frac{N_r}{N_y}, \quad
    \sigma^2_y = \sigma^2_h = \sigma^2, \quad
    \text{and}\quad
      \sigma^2_b = \sigma^2 \frac{N_b}{N_y} 
      \quad (b = dbr, nbl, nbt, nbb).
\end{eqnarray}
These weights are reported in Table \ref{tab:2d_gwf} for the considered noise variances $\sigma^2 = 1$, $\sigma^2 = 10^{-2}$, and $\sigma^2 = 10^{-4}$. 

\begin{figure}[!htb]
    \includegraphics[width=\textwidth]{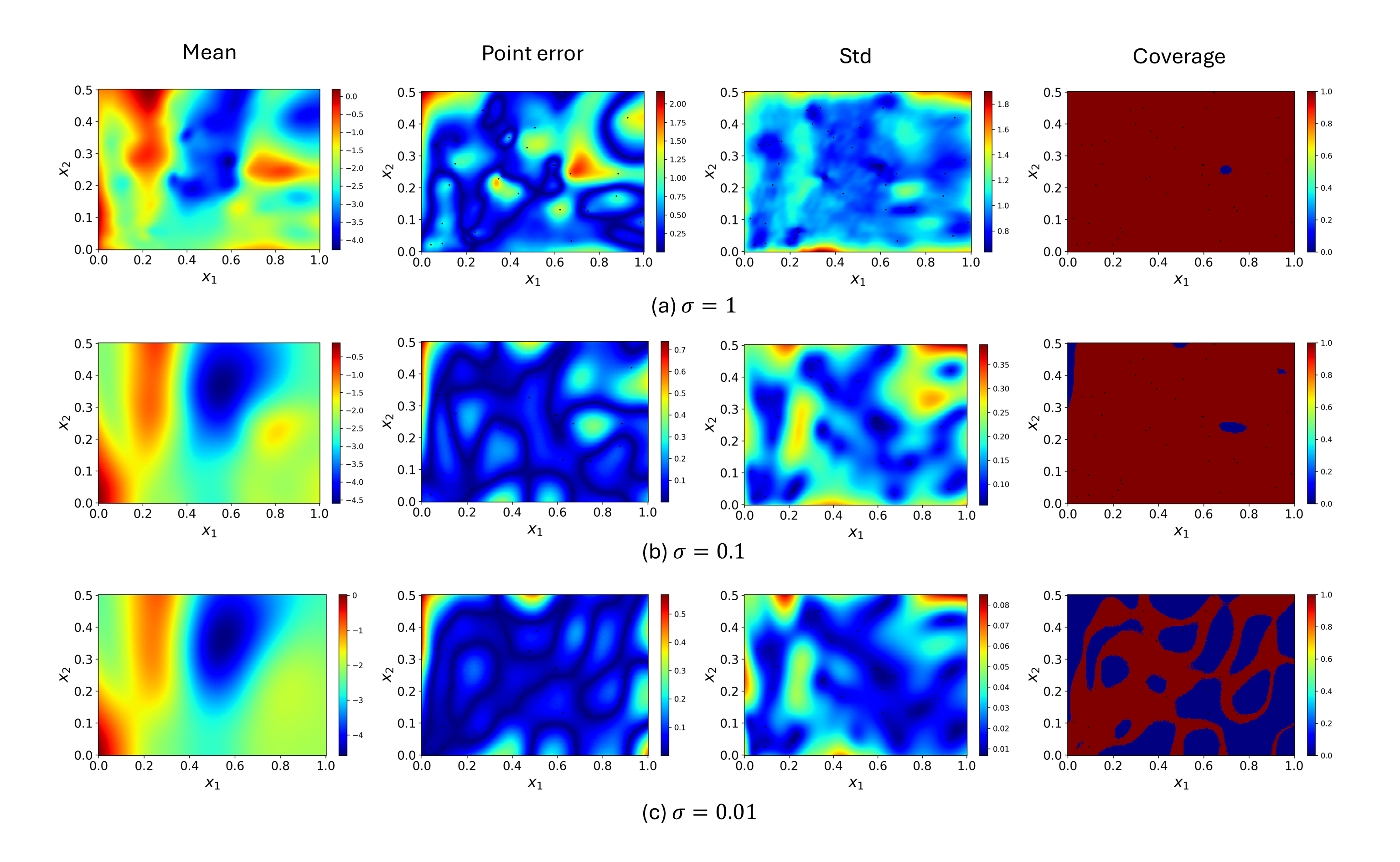}
    \caption{ Two-dimensional diffusion equation: rPINN predictions of $y(\mathbf{x})$ for (a) $\sigma = 1$, (b) $\sigma = 0.1$, and (c) $\sigma = 0.01$: (first column) the posterior mean estimates; (second column) point errors in the predicted mean $y$ with respect to the reference $y$ field; (third column) pointwise posterior standard deviation; and (fourth column) the coverage of the reference field with the $95\%$ credibility interval, where the zero and one values correspond to the reference field being outside and inside the credibility interval, respectively.}
    \label{fig:2d_gwf_rPINN_ypred}
\end{figure}

\begin{figure}[!htb]
    \includegraphics[width=\textwidth]{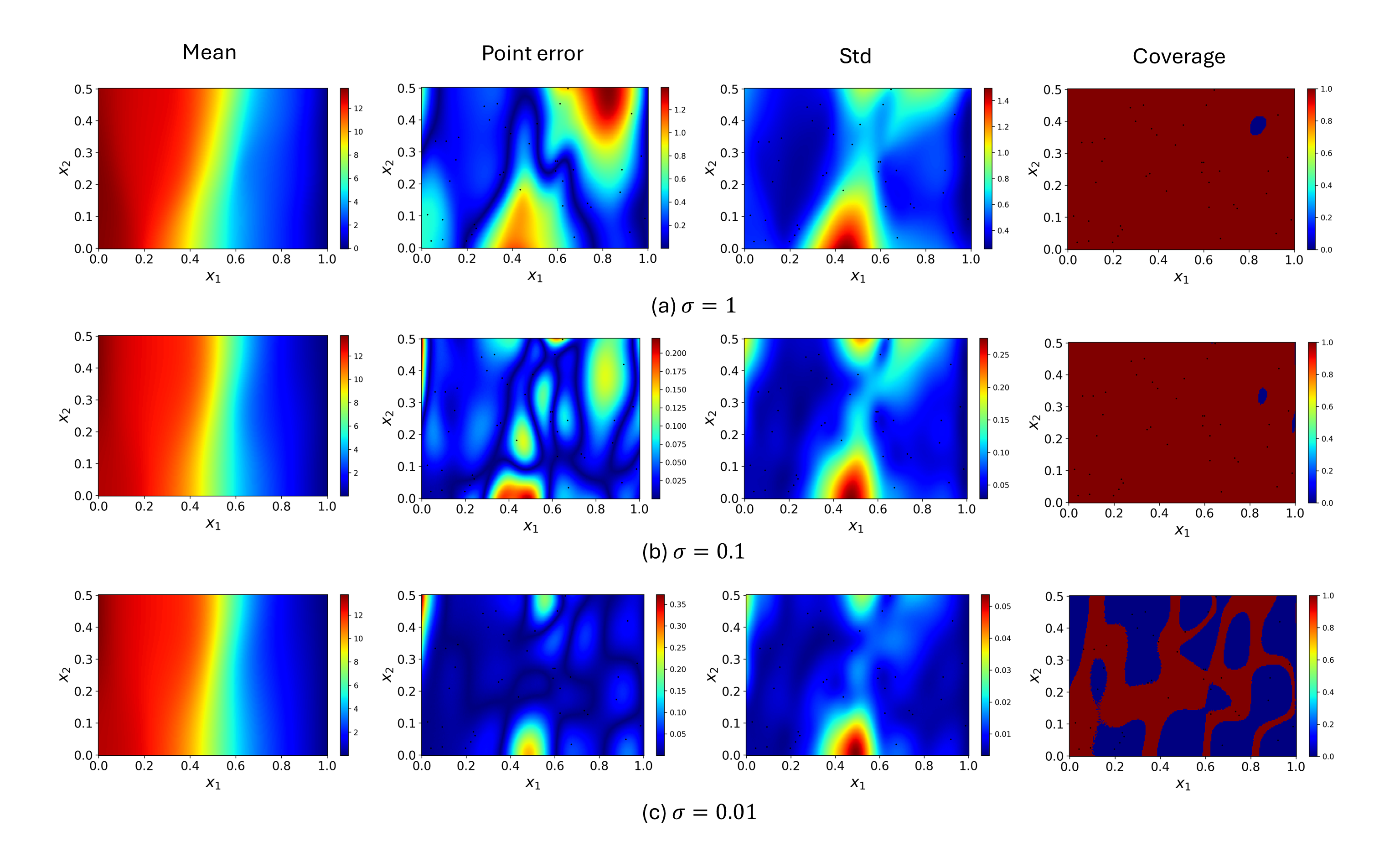}
    \caption{ Two-dimensional diffusion equation: posterior predictions of $h(\mathbf{x})$ given by rPINN for (a) $\sigma = 1$, (b) $\sigma = 0.1$, and (c) $\sigma = 0.01$.}
    \label{fig:2d_gwf_rPINN_hpred}
\end{figure}

Figures~\ref{fig:2d_gwf_rPINN_ypred} and~\ref{fig:2d_gwf_rPINN_hpred} show the rPINN estimate of the $\mu_y(\mathbf{x})$ and $\mu_h(\mathbf{x})$, the absolute pointwise differences $|\mu_y(\mathbf{x}) - y_{\text{ref}}(\mathbf{x})|$ and $|\mu_h(\mathbf{x}) - h_{\text{ref}}(\mathbf{x})|$, the posterior standard deviation at each grid point $\sigma_y(\mathbf{x})$ and $\sigma_h(\mathbf{x})$, and the coverage of $y_{\text{ref}}$ and $h_{\text{ref}}$ for all $\sigma^2$ values. 
Table~\ref{tab:2d_gwf} summarizes the relative $\ell^2$ and $\ell^{\infty}$ errors, LPP, and coverage in the rPINN and PINN-DE estimates of $y$ and $h$ posterior distributions. The rPINN and PINN-DE estimates of $\mu_y$ and $\mu_h$ are very close to those in PINN. However, for $\sigma = 1$ and $0.1$, PINN-DE predicts lower spatially averaged standard deviation, LPP, and coverage than PINN-DE, which indicates that PINN-DE underestimates uncertainty. 

A Bayesian model should become less informative with increasing variance, i.e., LPP should decrease with increasing $\sigma$. We find that $r\ell^2_y$ and $r\ell^2_h$ errors in rPINN increase with $\sigma$. Equally expected is that uncertainty in the predicted fields (i.e., $\sigma^2_y$ and $\sigma^2_h$) also increases with $\sigma$.  LPPs of $y$ and $h$ are highest for $\sigma=0.1$ and the lowest for $\sigma=1$. Large coverage (exceeding 98\%) indicates that rPINN provides informative posterior distributions for these values of $\sigma$. 
For  $\sigma = 0.01$,  the LPPs in the rPINN solutions are smaller than for $\sigma = 0.1$ and the coverage is relatively low. The quality of the posterior can be improved by modifying  Eq \eqref{eq:sigmap} for small $\sigma$ values as 
\begin{equation}\label{eq:sigmap_mod}
\sigma^2_p = (\sigma+\varepsilon)^2 \frac{\lambda_y}{N_y},  
\end{equation}
where $\varepsilon$ is a regularization parameter chosen to maximize LPP. 

In this problem, HMC fails to converge for all considered $\sigma$. For example, for the largest considered $\sigma$ (which yields the smoothest posterior that should be easiest for HMC to sample), the largest $\hat{R}$ computed from four HMC chains exceeds 20 (the indicator of convergence is $\hat{R}\approx 1$) and the estimated $\mu_y(\mathbf{x})$ and $\mu_h(\mathbf{x})$ fields significantly deviate from $y_{\text{ref}}(\mathbf{x})$ and $h_{\text{ref}}(\mathbf{x})$. The $r\ell^2_y$ errors for four chains are \num{3.4e-1}, \num{1.2e-0}, \num{2.4e-0},  and \num{1.4e-0}. The $r\ell^2_h$ errors for four chains are \num{5.8e-2}, \num{1.6e-0}, \num{2.1e-0}, \num{1.3e-0}. HMC also fails to converge for smaller values of $\sigma$. Therefore, HMC may not be suitable for this problem, where the required number of DNN parameters is ten times larger than in the one-dimensional Poisson problems. This underscores the advantages of rPINN in sampling high-dimensional neural network posterior distributions. 

\begin{table}[!htb]
\centering
\caption{Two-dimensional diffusion equation: $y$ and $h$ statistics predicted by rPINN and PINN-DE for $\sigma = 1, 0.1$ and $0.01$.}
\begin{adjustbox}{max width=\textwidth, max height=\textheight}
{\begin{tabular}{l l S[table-format=1.2e-1]S[table-format=1.2e-1]S[table-format=1.2e-1]S[table-format=6] S[table-format=2]}
\toprule
& \quad \textbf{Method} &\textbf{$r\ell^2$ error} & \textbf{$\ell^\infty$ error} & \textbf{Ave. Std} & \textbf{LPP} & \textbf{Coverage}  \\
\midrule
 \multicolumn{7}{l}{$\sigma = 1$: $\sigma_r = 3.5$, $\sigma_{dbr} = \sigma_{nbl} = 0.6$, $\sigma_{nbt} = \sigma_{nbb} = 0.9$, $\sigma_h = \sigma_y = 1$, $\sigma_p= 23.7$} \\
\hline
\multirow{2}{*}{$y(\mathbf{x})$} & rPINN & 2.6e-1 & 2.2 & 1.0 & -37254 &  99\%  \\
                                & PINN-DE & 2.8e-1 & 2.3 & 1.9e-1 & -320178 &  33\%  \\
\multirow{2}{*}{$h(\mathbf{x})$} & rPINN & 5.9e-2 & 1.4 & 5.4e-1 & 3978 &  99\%    \\    
                                & PINN-DE & 5.1e-2 & 1.4 & 7.4e-2 & -683133 &  24\%  \\
\hline
 \multicolumn{7}{l}{$\sigma = 0.1$: $\sigma_r = 0.35$, $\sigma_{dbr} = \sigma_{nbl} = 0.06$, $\sigma_{nbt} = \sigma_{nbb} = 0.09$, $\sigma_h = \sigma_y= 0.1$, $\sigma_p= 2.37$} \\
\hline
\multirow{2}{*}{$y(\mathbf{x})$} & rPINN  & 5.7e-2 & 7.4e-1 & 1.7e-1 & 80642 & 98\%    \\
                                & PINN-DE & 5.0e-2 & 5.9e-1 & 2.6e-2 & -230388 &  31\%  \\
\multirow{2}{*}{$h(\mathbf{x})$} & rPINN & 7.3e-3 & 3.7e-1 & 7.6e-2 & 138698 &  99\%  \\  
                                & PINN-DE & 6.5e-3 & 3.8e-1 & 9.4e-3 & -328889 &  26\%  \\
\hline
 \multicolumn{7}{l}{$\sigma = 0.01$: $\sigma_r = 0.04$, $\sigma_{dbr} = \sigma_{nbl} = 0.006$, $\sigma_{nbt} = \sigma_{nbb} = 0.009$, $\sigma_h = \sigma_y= 0.01$, $\sigma_p= 0.24$.} \\
\hline
\multirow{2}{*}{$y(\mathbf{x})$} & rPINN  &  4.0e-2 & 5.7e-1 & 2.5e-2 & -16959 &  49\%   \\
                                 & PINN-DE & 4.1e-2 & 5.7e-1 & 2.2e-2 & -107741 &  42\%  \\
\multirow{2}{*}{$h(\mathbf{x})$} & rPINN   & 5.7e-3 & 3.7e-1 &  1.0e-2 & 84341 & 45\%   \\
                                 & PINN-DE & 5.8e-3 & 3.8e-1 & 8.6e-3 & -44244 &  34\%  \\
\bottomrule
\end{tabular}
}
\end{adjustbox}
\label{tab:2d_gwf}
\end{table}

\section{Discussions and Conclusions}\label{sec:conclusions}
We present the rPINN method, an approximate Bayesian inference technique for sampling high-dimensional posterior distributions in PDE problems modeled with PINNs. The main idea of our approach is to introduce noises into the PINN loss function and approximate the posterior using samples obtained from the solutions of the resulting minimization problem for different noise realizations.

A distinct feature of the PINN model is the multi-term loss function, which translates into a multi-term likelihood in the BPINN model. To balance the contributions of different terms corresponding to various data types (PDE and boundary condition residuals, state variables, and parameters), we propose a weighted-likelihood BPINN formulation. This formulation is consistent with the weight selection approaches in the PINN method. Furthermore, we proposed an empirical expression for selecting the prior variance of the DNN weights as a function of the measurement noise variance. 

The robustness of the rPINN method was demonstrated via comparison with BPINN-HMC, BPINN-SVGD, and deep ensemble PINN methods for linear and non-linear inverse Poisson equations and the inverse diffusion equation with space-dependent diffusion coefficient. The rPINN method yielded highly informative predictive posterior distributions with relatively small $r\ell_2$ errors and large LPPs and coverage for the considered problems. PINN-DE underestimated the total uncertainty, resulting in significantly smaller variances and LPPs.

For the linear Poisson problem, we observed good agreement between rPINN, BPINN-HMC, and BPINN-SVGD posterior distributions of state variables and parameters. However, we found that multiple HMC chains do not fully mix, which indicates the lack of convergence in the posterior space of the DNN parameters.  We also demonstrated that the weighted-likelihood BPINN formulation leads to posterior distributions that become more informative with the increasing number of observations. On the other hand, the original ``unweighted'' likelihood formulation produced less informative posteriors for the larger number of measurements.  For the non-linear Poisson problem and diffusion equation, the BPINN-HMC and BPINN-SVGD failed to converge, yielding poor predictions of the state variables. We attribute the failure of these methods to the complex shape of the posterior distributions of the PINN parameters with multiple poorly connected modes. 
 
Our work demonstrates that rPINN can be an attractive alternative to HMC, BPINN-SVGD, and the deep ensemble methods for sampling high-dimensional posterior distributions of the PINN parameters and quantifying uncertainty in inverse PDE solutions. 

\bibliographystyle{elsarticle-num}

\appendix

\section{Hamiltonian Monte Carlo (HMC)}\label{sec:HMC}
HMC samples the posterior by expanding the parameter space $\boldsymbol\theta$ to an augmented phase space $(\boldsymbol\theta, \boldsymbol\rho)$ by introducing an auxiliary momentum variable $\boldsymbol\rho$, which is of the same size as $\boldsymbol\theta$. During each iteration, a new point $(\boldsymbol\theta^*, \boldsymbol\rho^*)$ in the phase space is proposed by simulating the trajectory of the Hamiltonian dynamics starting from the initial value of $(\boldsymbol\theta_0, \boldsymbol\rho_0)$:
\begin{eqnarray}\label{eq:hamiltonian_dynamics}
    \frac{d\boldsymbol\theta}{dt} &=& -\frac{\partial \mathrm{H}}{\partial \boldsymbol\rho} \\
    \frac{d\boldsymbol\rho}{dt} &=& +\frac{\partial \mathrm{H}}{\partial \boldsymbol\theta},
\end{eqnarray}
where $\mathrm{H}(\boldsymbol\theta, \boldsymbol\rho)$ is the Hamiltonian, which includes the potential energy function $\mathrm{U}(\boldsymbol\theta)$ and the kinetic energy function $\mathrm{K}(\boldsymbol\rho)$. In BPINN-HMC, we use the energy function Eq \eqref{eq:energy_loss} (the negative log density of the posterior), as $\mathrm{U}(\boldsymbol\theta)$. The kinetic energy is chosen as the negative log density of a zero-mean multivariate Gaussian with a covariance matrix $M$, also called the mass matrix. A Metropolis algorithm is employed for rejecting the proposed state to address deviations from the exact trajectory due to numerical integration. The proposed state is accepted with the probability $\min[1, \exp(-\mathrm{H}(\boldsymbol\theta^*, \boldsymbol\rho^*) + \mathrm{H}(\boldsymbol\theta_0, \boldsymbol\rho_0))]$. The new sample from the posterior is obtained by projecting Hamiltonian back to the sample space $\boldsymbol\theta$. In this work, we use the No-U-Turn Sampler (NUTS) \cite{hoffman2014NUTS}, an extension of HMC, which automatically tunes the number of integration steps in HMC. Furthermore, we employ the dual averaging step-size adaptation algorithm~\cite{hoffman2014NUTS} to automatically determine the optimal HMC step size for maintaining a desired acceptance rate (we set the acceptance rate to $75\%$).  Further details on HMC can be found in \cite{neal2011mcmc,betancourt2017conceptual}. HMC is known to suffer from the "curse of dimensionality."  

\subsection{Convergence Diagnostics of BPINN-HMC}
Analyzing the HMC convergence to the true posterior distribution is important to ensure the reliability of HMC predictions. We use Gelman and Rubin's $\hat{R}$ potential scale reduction factor \cite{gelman1992rhat} as a quantitative convergence indicator.  The $\hat{R}$ factor is computed for each parameter, and its value close to one indicates the convergence for this dimension. We compute $\hat{R}$ for all parameters and plot the histograms of $\hat{R}$ values. Furthermore, we provide the trace plot for each chain showing samples of two DNN parameters (with the highest and lowest $\hat{R}$ values) at each iteration after the burn-in phase.  We also show the negative log probability computed at each step after the burn-in phase for all HMC chains. If the negative log probabilities computed from different chains oscillate within different intervals, it indicates the lack of mixing (chains sample around different modes). However, having HMC chains with  ``mixed'' log posterior densities is not sufficient for the HMC convergence, because different modes in the DNN parameter space can provide similar log probabilities. 

\begin{figure}[!htb]
    \begin{adjustbox}{max width=\textwidth, max height=0.92\textheight}
    \centering
    \includegraphics[width=0.92\textwidth]{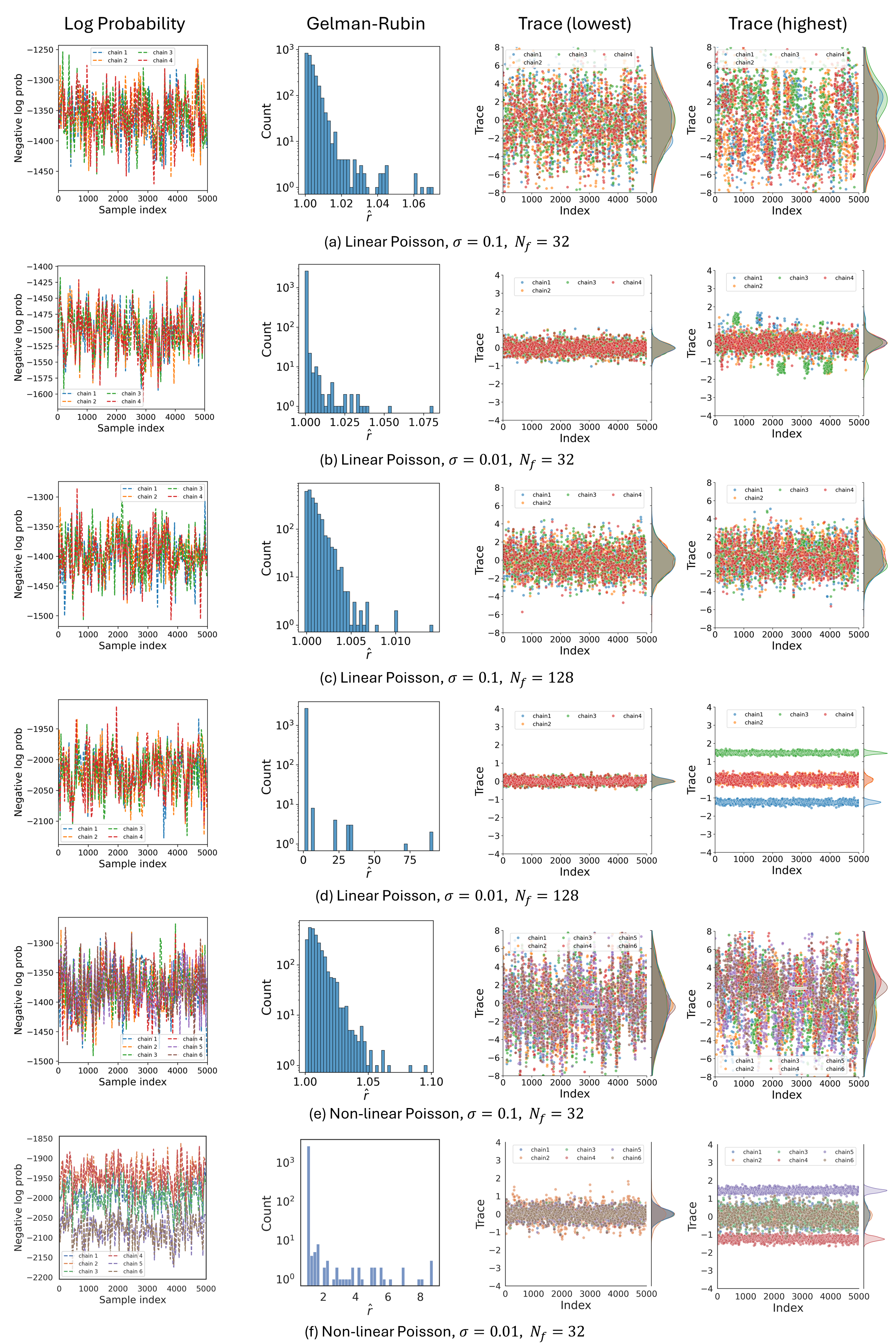}
    \end{adjustbox}
    \caption{ The HMC convergence analysis for the linear and non-linear Poisson equations: (first column) the log posterior density computed from different chains, (second column) the histogram of $\hat{R}$, and (third and fourth columns) trace plots for the parameters with $\hat{R}$ closest to one and the largest $\hat{R}$, respectively. }
    \label{fig:HMC_convergence}
\end{figure}

Figure \ref{fig:HMC_convergence} shows the negative log probability, the histogram of $\hat{R}$, and traces for the liner and non-linear inverse Poisson HMC solutions for several values of $N_f$ and $\sigma$. For the linear problem, HMC converges for all DNN parameters, except the case when $N_f = 128$ and $\sigma = 0.01$. 
For the non-linear problem, HMC does not converge for any of the considered cases. Below we investigate the convergence behavior of HMC for the linear and non-linear Poisson equations in more detail. 

\subsection{Linear Poisson Equation}
We run four chains for the linear Poisson equation. For $N_f = 32$ and $128$ and $\sigma = 0.1$, we find that $50000$ burn-in iterations are sufficient for HMC to converge. On the other hand, for $N_f = 128$ and $\sigma = 0.01$, we find that the chains do not mix even after $50000$ burn-in iterations. Large $\hat{R}$ values are obtained for several DNN parameters, and the trace plot indicates poor mixing. Despite the non-convergence, different HMC chains still produce similar means, variances, and LPPs. 

The loss landscape of PINN is highly non-convex with multiple local minima. This translates into a highly complex BPINN posterior density with numerous local modes. For $\sigma = 0.1$, each chain samples multiple modes. For $\sigma = 0.01$ and $N_f = 128$, each chain only samples certain modes. We assume this happens because different modes are better connected for larger $\sigma$.  
%
Despite poor mixing in the parameter space $\boldsymbol\theta$, the HMC chains produce consistent results in the function space of $f$. 
Better mixing in the function space than the parameter space was also reported in ~\cite{izmailov2021bayesian}. 
For the linear Poisson equation, consistent results in the function space of $f$ translate into consistent predictions in the function space of $u$. 

\subsection{Non-Linear Poisson Equation}
Here, we initialize six chains and set burn-in iterations to $100000$.  For $\sigma = 0.1$, the majority of the DNN parameters have $\hat{R}$ values close to 1. However,  several parameters have large $\hat{R}$ values. Lack of convergence along the corresponding dimensions in the probability space causes the overall failure of the HMC method to converge to the correct $u$ posterior. For $\sigma = 0.01$, the lack of convergence is even more pronounced as more parameters have large $\hat{R}$ values. The trace plot for the worst mixing dimension illustrates three distinct marginal distributions.
This suggests that chains sample the posterior distribution around different modes. To confirm the multimodality, for $\sigma = 0.01$ we plot the projection of the posterior log density on a two-dimensional subspace of the DNN parameters constructed as \cite{garipov2018loss, izmailov2021bayesian}:
\begin{eqnarray}\label{eq:S_space}
    \mathcal{S}=\left\{(a,b) \mid \boldsymbol\theta = \boldsymbol\theta_1 \cdot a + \boldsymbol\theta_2 \cdot b + \boldsymbol\theta_3 \cdot(1-a-b)\right\}.
\end{eqnarray}
Here, $\boldsymbol\theta_1$, $\boldsymbol\theta_2$, and $\boldsymbol\theta_3$ are three parameter vectors selected to construct the subspace, and $(a, b)$ are the coordinates of the arbitrary parameter vector $\boldsymbol\theta$ in the subspace $\mathcal{S}$. Specifically, $\boldsymbol\theta_1 - \boldsymbol\theta_2$ and $\boldsymbol\theta_1 - \boldsymbol\theta_3$ serve as independent basis vectors, and the Gram-Schmidt process is performed to make them orthonormal. We select two sets of the subspace-spanning vectors.   In the first set, three vectors are selected as the last samples from three different HMC chains (chains 1, 3, and 6) sampling distinctly different $u$ distributions. In the second set, three vectors are selected from one HMC chain, where $\boldsymbol\theta_1$, $\boldsymbol\theta_2$, and $\boldsymbol\theta_3$ are given by the first (after burn-in), the middle, and the last samples, respectively, in the given chain. 

\begin{figure}[!htb]
	\centering
	\subfloat[] {\includegraphics[angle=0,width=0.33\textwidth]{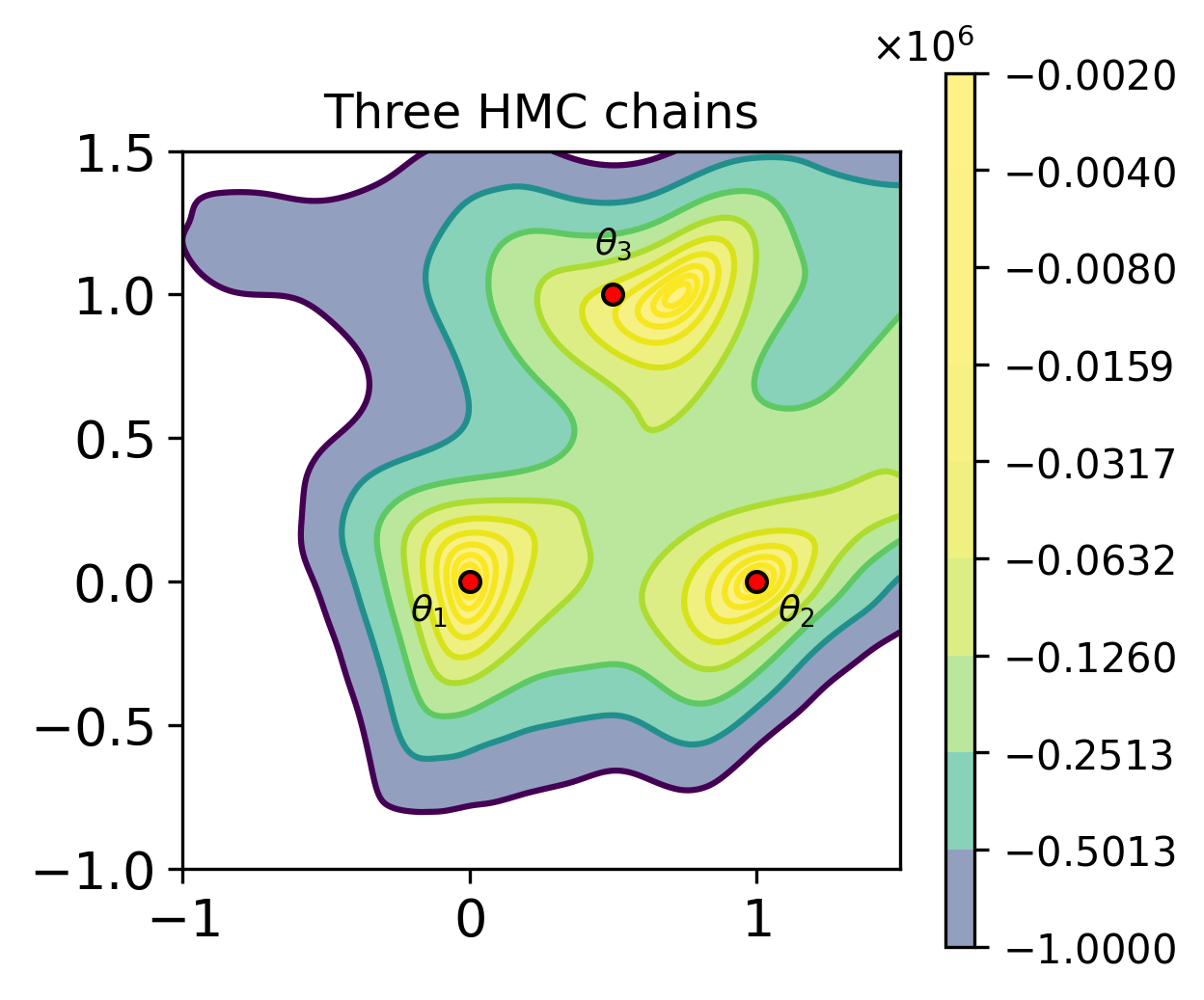}}
	\subfloat[] {\includegraphics[angle=0,width=0.33\textwidth]{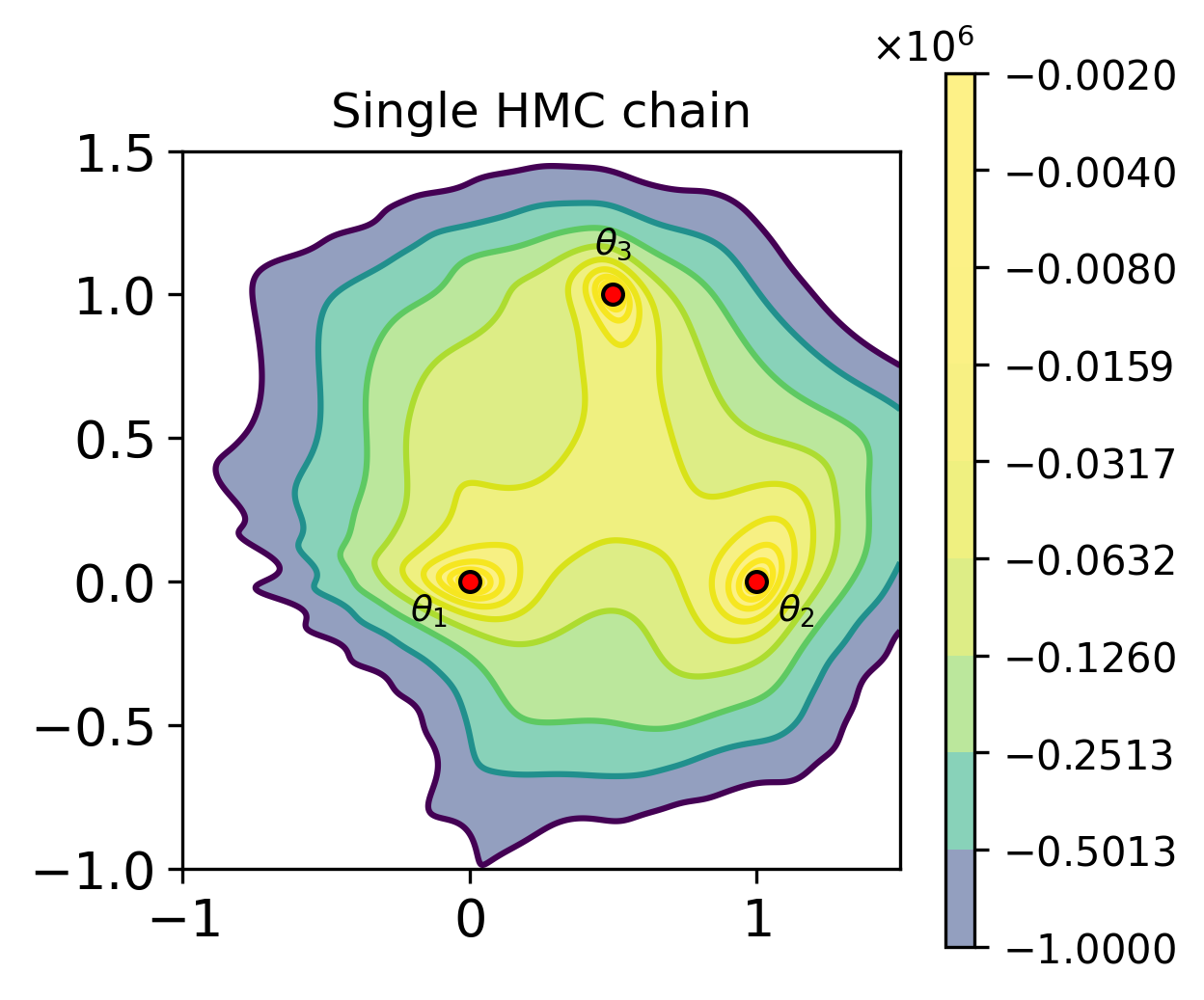}}
    \subfloat[] {\includegraphics[angle=0,width=0.28\textwidth]{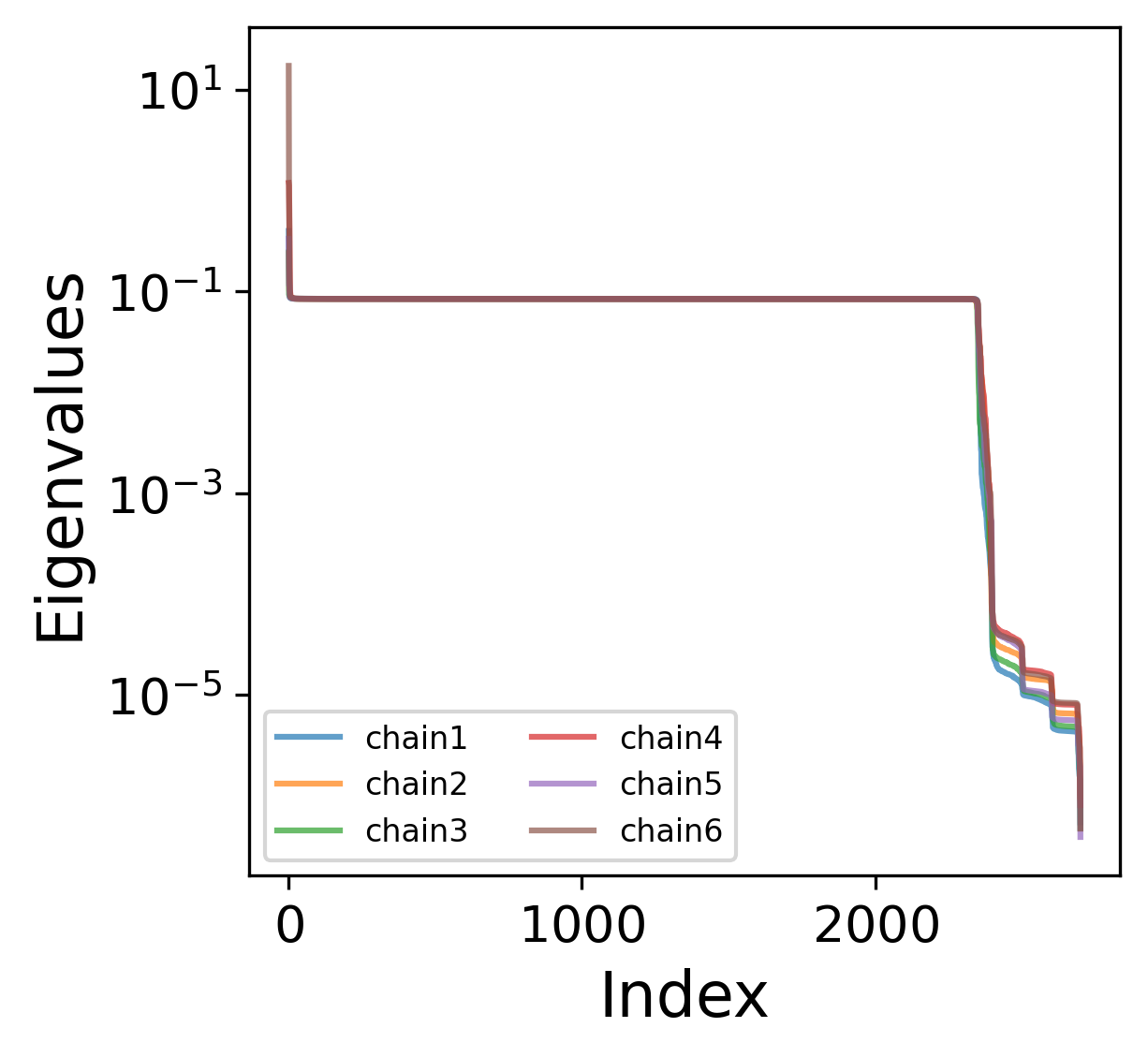}}
	\caption{ Projection of the posterior log density on the two-dimensional subspace  spanned by three
HMC samples from (a) three independent chains (chains 1, 3, and 6), and (b) the same chain. Panel c shows the eigenvalue spectrum of the Hessian evaluated at the modes of the the distributions sampled by six chains. }
	\label{fig:HMC_posterior_visialization}
\end{figure}

Figure~\ref{fig:HMC_posterior_visialization} shows the two-dimensional log posterior density plot for the two choices of $\boldsymbol\theta_1$, $\boldsymbol\theta_2$, and $\boldsymbol\theta_3$. Three modes exist for both sets of $\boldsymbol\theta_i$ vectors. However, the modes are more pronounced (and, therefore, more isolated) for $\boldsymbol\theta_i$ drawn from three chains. 
This indicates that these three modes, which correspond to three HMC chains, are more isolated. 
Additionally, in Figure~\ref{fig:HMC_posterior_visialization}(c), we show the eigenvalues of the Hessian of the log posterior density with respect to the DNN parameters evaluated at the last sample of the six chains. 
The variability in the eigenvalue spectrum tail also suggests that difference chains are moving around different modes of the posterior distribution. 

\section{Stein Variational Gradient Descent (SVGD)}\label{sec:SVGD}
SVGD is a non-parametric variational inference approach for approximating a posterior distribution without assuming its specific form. SVGD moves a set of particles $\{ \boldsymbol\theta_i \}_{i = 1}^{N_{ens}}$ around the parameter space guided by the gradient of the Kullback–Leibler (KL) divergence to minimize the KL divergence between the particle distribution and the target posterior.  
The main challenges in SVGD include sensitivity to the choice of kernel, computational costs in high-dimensional settings, and difficulty in capturing multi-modal distributions. 
To compute the KL divergence, we utilize the kernelized Stein operator within the unit ball of a reproducing kernel Hilbert space. More concretely, at iteration level $t$, the update equations mimic a gradient descent update:
\begin{eqnarray}
    \boldsymbol\theta_i^{t+1} = \boldsymbol\theta_i^{t} + \alpha^t \phi(\boldsymbol\theta_i^{t})
\end{eqnarray}
where $\alpha^t$ is the step size at iteration $t$, and the ``update function'' is given as the sum of the gradient of the log posterior and a repulsion term derived from a kernel function:
\begin{eqnarray}
    \phi(\boldsymbol\theta)=\frac{1}{n} \sum_{j=1}^n[k\left(\boldsymbol\theta_j^t, \boldsymbol\theta\right) \underbrace{\nabla_{\boldsymbol\theta_j^t}\left(\log p\left(\boldsymbol\theta_j^t\right)+\log p\left(\mathcal{D}\mid \boldsymbol\theta_j^t\right)\right)}_{\text {gradient }}+\underbrace{\nabla_{\boldsymbol\theta_j^t} k\left(\boldsymbol\theta_j^t, \boldsymbol\theta\right)}_{\text {repulsive force }}]
\end{eqnarray}
where $k(\cdot, \cdot): \mathbb{R}^{N_\theta} \times \mathbb{R}^{N_\theta} \rightarrow \mathbb{R}$ is the kernel function, $p(\boldsymbol\theta)$ is the prior and $p(\mathcal{D} | \boldsymbol\theta)$ is the likelihood. In BPINN-SVGD, we use the gradient of Eqn \eqref{eq:energy_loss} to drive particles. The gradient is computed by automatic differentiation. 
In this work, we use the Adam optimizer to update the positions of the particles with the learning rate set to \num{1e-3}. The radial basis kernel is used with the correlation length determined during sampling by the ``median trick''~\cite{liu2019stein}.

\end{document}